\documentclass[a4paper,fleqn]{cas-dc}
\graphicspath{ {./figures/} }
\usepackage{amssymb}
\usepackage{multirow}
\usepackage{algorithmic}
\usepackage{array}

\usepackage{caption}

\usepackage{textcomp}
\usepackage{stfloats}
\usepackage{url}
\usepackage{verbatim}
\usepackage{graphicx}
\usepackage{epstopdf}
\usepackage{booktabs, siunitx, dcolumn}
\usepackage{silence}
\usepackage{bm}
\usepackage[round,authoryear]{natbib}
\usepackage{subfigure}
\usepackage{amsmath,amsfonts}
\usepackage[ruled]{algorithm2e}
\usepackage{soul}
\usepackage{bm}
\usepackage{color, xcolor}
\usepackage{times}
\usepackage{epsfig}
\usepackage{tabulary}
\usepackage{tabularx}
\usepackage{enumitem}
\usepackage{comment}
\usepackage{amssymb,amsfonts} % define this before the line numbering.
\usepackage{nicefrac}       % compact symbols for 1/2, etc.
\usepackage{microtype}      % microtypography
\usepackage{wrapfig}
\usepackage[export]{adjustbox}
\usepackage[percent]{overpic}
\usepackage{pifont}

\newcommand*{\eg}{\textit{e}.\textit{g}.\@\xspace}
\newcommand*{\ie}{\textit{i}.\textit{e}.\@\xspace}

\newcommand{\cmark}{\ding{51}}%

\newcolumntype{L}[1]{>{\raggedright\let\newline\\\arraybackslash\hspace{0pt}}m{#1}}
\newcolumntype{R}[1]{>{\raggedleft\let\newline\\\arraybackslash\hspace{0pt}}m{#1}}
\newcolumntype{C}[1]{>{\centering\let\newline\\\arraybackslash\hspace{0pt}}m{#1}}

\soulregister \citep7
\soulregister \citepp7
\soulregister\ref7

\def\tsc#1{\csdef{#1}{\textsc{\lowercase{#1}}\xspace}}
\tsc{WGM}
\tsc{QE}
\begin{document}
\let\WriteBookmarks\relax
\def\floatpagepagefraction{1}
\def\textpagefraction{.001}

%\shorttitle{C. Zhang et~al./ Expert Systems with Applications}
%\shortauthors{zcy et~al.}

% Main title of the paper
\title [mode = title]{3DPillars: Pillar-Based Two-Stage 3D Object Detection}

% Title footnote mark
% eg: \tnotemark[1]
% \tnotemark[<tnote number>] 

% Title footnote 1.
% eg: \tnotetext[1]{Title footnote text}
% \tnotetext[<tnote number>]{<tnote text>} 

\author[1]{Jongyoun Noh}
\ead{jyoun.noh@samsung.com}

\author[2]{Junghyup Lee}
\ead{junghyup.lee@samsung.com}

\author[3]{Hyekang Park}
\ead{hyekang.park@yonsei.ac.kr}

\author[3]{Bumsub Ham}
\cormark[1]
\ead{bumsub.ham@yonsei.ac.kr}

\affiliation[1]{organization={Samsung Electronics},
            city={Suwon-si, Gyeonggi-do},
            postcode={16677},
            country={South Korea}}
\affiliation[2]{organization={Samsung Research},
			city={Seoul},
			postcode={06765},
			country={South Korea}}
\affiliation[3]{organization={School of Electrical and Electronic Engineering, Yonsei University},
            city={Seoul},
            postcode={03722},
            country={South Korea}}
\cortext[1]{Corresponding author.}

\begin{abstract}
  PointPillars is the fastest 3D object detector that exploits pseudo image representations to encode features for 3D objects in a scene. Albeit efficient, PointPillars is typically outperformed by state-of-the-art 3D detection methods due to the following limitations: 1)~The pseudo image representations fail to preserve precise 3D structures, and 2)~they make it difficult to adopt a two-stage detection pipeline using 3D object proposals that typically shows better performance than a single-stage approach. We introduce in this paper the first two-stage 3D detection framework exploiting pseudo image representations, narrowing the performance gaps between PointPillars and state-of-the-art methods, while retaining its efficiency. Our framework consists of two novel components that overcome the aforementioned limitations of PointPillars: First, we introduce a new CNN architecture, dubbed 3DPillars, that enables learning 3D voxel-based features from the pseudo image representation efficiently using 2D convolutions. The basic idea behind 3DPillars is that 3D features from voxels can be viewed as a stack of pseudo images. To implement this idea, we propose a separable voxel feature module that extracts voxel-based features without using 3D convolutions. Second, we introduce an RoI head with a sparse scene context feature module that aggregates multi-scale features from 3DPillars to obtain a sparse scene feature. This enables adopting a two-stage pipeline effectively, and fully leveraging contextual information of a scene to refine 3D object proposals. Experimental results on the KITTI and Waymo Open datasets demonstrate the effectiveness and efficiency of our approach, achieving a good compromise in terms of speed and accuracy. 
\end{abstract}

\begin{keywords}
	LIDAR \sep Point clouds \sep 3D object detection \sep Pillar-based learning \sep Autonomous driving
\end{keywords}

\maketitle

\begin{figure*}[t]
	% \captionsetup{font={footnotesize}}
	   \centering
		  \includegraphics[width=1.0\linewidth]{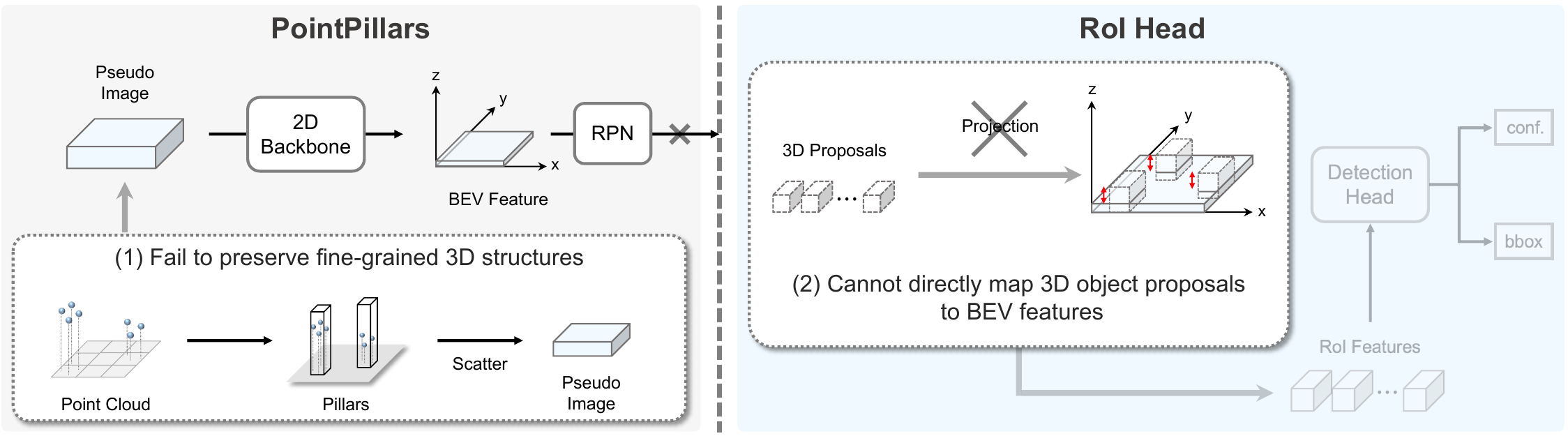}
  \caption{Limitations of PointPillars~\citep{lang2019pointpillars}. A pseudo image representation in PointPillars allows to exploit 2D convolutions for efficient 3D object detection. Left: The pseudo image representation however collapses point clouds in vertical pillars, and thus PointPillars does not preserve precise 3D structures of a scene. Right: 3D object proposals cannot be projected onto BEV features directly, suggesting that two-stage detection frameworks using 3D proposals could not be easily adopted to PointPillars.}
   \label{fig:teaser}
  \vspace{-0.4cm}
\end{figure*}

\section{Introduction}\label{sec:intro}
3D object detection aims at localizing 3D objects in a scene while predicting their class labels. As a fundamental task for 3D scene understanding, it serves as a key component for many applications, including autonomous driving and robotics. Recently, LiDAR-based 3D object detection has attracted a lot of attention that exploits raw point clouds acquired from a LiDAR sensor. The main challenge of LiDAR-based 3D object detection is how to learn discriminative 3D representations with sparse and irregular point clouds. Point-based methods~\citep{qi2018frustum,shi2019pointrcnn,wang2019frustum,yang2019std,yang20203dssd,shi2020pv,shi2020point,li2024ws,wu2024dccn} extract point-wise features directly using PointNet or its variants~\citep{qi2017pointnet,qi2017pointnet++}, which however usually require computationally expensive operations to aggregate local information. They typically sample point clouds to reduce the number of large-scale point clouds, limiting the performance of 3D object detection. An alternative approach is to transform point clouds to structured grid representations, such as 3D voxels~\citep{he2020structure,deng2021voxel,zhou2018voxelnet,yan2018second,yin2021center,hu2022density,shi2023pv,guang2024rpea} or~a bird's-eye view~(BEV) represetations~\citep{wang2020pillar,lang2019pointpillars,yang2018pixor,ye2020hvnet,noh2021hvpr,liu2020tanet, shi2022pillarnet}, rather than directly  extracting 3D features from the point clouds. Voxel-based methods~\citep{he2020structure,deng2021voxel,zhou2018voxelnet,yan2018second,yin2021center,hu2022density,shi2023pv} divide point clouds into a 3D voxel grid to extract voxel-wise features using 3D convolutional neural networks~(CNNs). BEV-based methods~\citep{lang2019pointpillars,yang2018pixor,ye2020hvnet,shi2022pillarnet,simony2018complex, wang2020pillar} obtain a compact 3D representation in a form of an image by projecting point clouds~\citep{yang2018pixor,simony2018complex} or gathering points in vertical pillars~\citep{wang2020pillar,lang2019pointpillars,liu2020tanet,ye2020hvnet,noh2021hvpr,shi2022pillarnet}, and then exploit standard 2D CNNs for feature extraction. Generally, BEV-based methods are more efficient in terms of speed and the number of parameters than the voxel-based ones, benefitting from 2D CNNs, but a structural loss is inevitable when projecting 3D point clouds into a 2D space. On the one hand, voxel-based methods preserve fine-grained 3D structures well, and thus they provide better detection results than BEV-based ones. Exploiting fine voxel grids and 3D CNNs, on the other hand, is computationally demanding.

PointPillars~\citep{lang2019pointpillars} is the most representative BEV-based method that introduces a pseudo image representation for 3D object detection. Although it is efficient, the detection accuracy is much lower than state-of-the-art 3D detectors~\citep{deng2021voxel,shi2023pv,miao2021pvgnet,mao2021pyramid,hu2022density}. There are mainly two reasons as follows:~(Fig.\ref{fig:teaser}): First, PointPillars does not encode fine-grained 3D structures, since it rearranges point clouds in vertical pillars. Second, a two-stage framework, which is commonly adopted by the state-of-the-art approaches~\citep{shi2023pv,deng2021voxel,miao2021pvgnet,mao2021pyramid,hu2022density}, is not applicable to PointPillars, as 3D object proposals in the two-stage framework would not be mapped directly onto BEV features, obtained from 2D CNNs. 
We analyze PointPillars in detail because it is a foundational and widely adopted BEV-based method; however, its structural limitations are not unique. Several of its variants and follow-up methods~\citep{wang2020pillar,liu2020tanet,noh2021hvpr,shi2022pillarnet} share the same limitations in representing fine-grained 3D structures and in adopting a single-stage detection framework.

We present in this paper a novel two-stage 3D object detection framework that bridges the performance gaps between pillar-based 3D object detectors~\citep{lang2019pointpillars,wang2020pillar,liu2020tanet,noh2021hvpr, shi2022pillarnet} and state-of-the-art methods. To this end, we introduce two components that overcome the aforementioned limitations of PointPillars~\citep{lang2019pointpillars}. First, we propose a new CNN architecture, dubbed 3DPillars, that leverages a set of pseudo images to learn voxel-based features, while encoding 3D structures of a scene. The basic idea behind 3DPillars is that 3D features from voxels can be viewed as a stack of pseudo images along each axis,~\ie, X, Y and Z~(Fig.\ref{fig:svfm}). That is, we can reconstruct the 3D features by stacking pseudo images along the Z-axis or other view-dependent ones along the X- or Y-axes. Note that we also consider pseudo image representations along X- and Y-axes to encode view-specific features in contrast to PointPillars exploiting BEV representations along the Z-axis only. To implement this idea, we introduce a separable voxel feature module~(SVFM) consisting of a series of 2D convolutional layers. Specifically, instead of using 3D convolutions, SVFM exploits standard 2D convolutions to abstract view-specific features for each pseudo image. This allows the model to preserve 3D structures without relying on expensive 3D convolutions, while enabling efficient voxel feature learning.
% It is worth to note that SVFM is similar to the depth-wise separable convolutions in that both factorize the convolutional operations. SVFM can be thought of as a general version of the separable convolutions, since the former is defined using a 4D tensor, while the latter applies to a 3D tensor.
Second, we introduce an RoI head with a sparse scene context feature module~(S$^{2}$CFM).
State-of-the-art 3D detectors~\citep{deng2021voxel,shi2023pv,mao2021pyramid,hu2022density} typically adopt a two-stage framework. In the second stage, they apply 3D RoI pooling methods on feature volumes from a backbone network to preserve precise structural information in 3D object proposals, which is crucial for the detection performance. The 3D RoI pooling methods are however designed to operate with sparse 3D CNNs, suggesting that we could not directly adapt them to 3D detection methods using standard 2D convolutions, including PointPillars and 3DPillars. Moreover, the pooling methods do not fully leverage contextual information learned from backbone networks, and they do not consider global scene contexts to refine RoIs. To alleviate these problems, S$^{2}$CFM aggregates multi-scale features into a sparse scene feature, encoding richer scene contexts into RoI features effectively and efficiently, and boosting the 3D detection performance significantly. S$^{2}$CFM further refines the RoI features using global contexts of a 3D scene, which is particularly effective to localize small objects. Extensive experimental results on standard benchmarks demonstrate the effectiveness and efficiency of our two-stage approach that overcomes the limitations of the pillar-based method. We will make our source code and trained models available online. The main contributions of our work can be summarized as follows:
\vspace{-0.1cm}

% We will release our source code and trained models online.

% \todo{The separable convolutions are typically exploited in lightweight network architectures in order to reduce the computational complexity.} 
% Each RoI feature is pooled from heuristically assigned features at certain backbone levels, which ignores context features from other levels. \todo{Also, to refine each object proposal, the features inside and very close to the proposal are used only without considering global scene contexts.this pooling method, designed for sparse convolutional backbone network, cannot be easily embedded to our framework. Also, as discussed in section 1, it cannot fully consider the contextual information learned from the network when refining the proposals. We propose S$^{2}$CFM and plug it in the RoI head to overcome the aforementioned problems, successfully settling the 2-stage pipeline in our framework.
% \vspace{-0.05cm}

\begin{itemize}
   \item We introduce a novel two-stage framework for 3D object detection exploiting pseudo image representations. To this end, we propose 3DPillars that extracts 3D voxel-based features efficiently using pseudo image representations without exploiting 3D convolutions.
   \item We introduce a new RoI head with S$^{2}$CFM, enabling employing a two-stage pipeline to 3DPillars, while considering rich scene contexts to refine 3D proposals.   
   \item We demonstrate that our approach gives a better compromise in terms of speed and accuracy, compared to the state of the art. We achieve competitive performance on both KITTI~\citep{geiger2012we}, and Waymo Open~\citep{sun2020scalability} datasets.
\end{itemize}
\vspace{-0.4cm}

\section{Related work}
In this section, we briefly describe represented works pertinent to ours. We classify LiDAR-based 3D object detection methods into three categories, according to types of 3D representations.
\vspace{-0.2cm}

\subsection{BEV-based approach}
CNNs have shown significant breakthroughs in 2D object detection. To adopt 2D detection techniques to 3D object detection directly, BEV-based approaches transform irregular point clouds into a form of a 2D image~\citep{wang2020pillar,lang2019pointpillars,yang2018pixor,ye2020hvnet,noh2021hvpr,liu2020tanet, shi2022pillarnet,du2020associate}. Early works project raw point clouds onto a discrete X-Y grid to obtain a BEV representation, and exploit 2D CNNs to generate 3D bounding boxes~\citep{yang2018pixor,simony2018complex}. PointPillars~\citep{lang2019pointpillars} introduces a new BEV representation,~pillars, by rearranging point clouds in vertical pillars, which preserves 3D structures better than the early BEV methods~\citep{yang2018pixor,simony2018complex}. Many variants to PointPillars are proposed to obtain better 3D representations~\citep{du2020associate,ye2020hvnet,noh2021hvpr,liu2020tanet,wang2020pillar,shi2022pillarnet}. They enhance the pillar-based representation by exploiting other types of features, such as multi-scale features~\citep{ye2020hvnet}, multi-view features~\citep{wang2020pillar}, and point-based features~\citep{noh2021hvpr}, or by introducing an attention mechanism~\citep{liu2020tanet}. Recently, PillarNet~\citep{shi2022pillarnet} proposes to use a sparse convolutional encoder network, improving the detection performance of PointPillars~\citep{lang2019pointpillars}. Recently, BEVSpread~\citep{wang2024bevspread} propagates image features to multiple BEV grids to improve the voxel pooling process, thereby alleviating position approximation errors in the BEV-based approaches. The pillar representation allows to obtain BEV features using 2D CNNs, but representing point clouds as pseudo images losses 3D structures of a scene. Our approach also uses pseudo images to represent point clouds, taking advantages of 2D CNNs in terms of efficiency. In contrast to other pillar-based methods, we exploit a set of pseudo images, and stack them along each axis of a 3D scene, retaining fine-grained 3D structures of point clouds. 

%We also leverage a two-stage pipeline in the pillar-based framework, improving the detection performance significantly.        

\subsection{Voxel-based approach}
Voxel-based approaches divide irregular point clouds into voxel grids defined on a 3D space, and apply 3D convolutions to extract voxel-wise features~\citep{he2020structure,deng2021voxel,zhou2018voxelnet,yan2018second,yin2021center,hu2022density,shi2023pv}. A pioneer work of~\citep{zhou2018voxelnet} exploits PointNet~\citep{qi2017pointnet} to obtain discriminative voxel representations. SECOND~\citep{yan2018second} introduces sparse 3D convolutions for an efficient 3D voxel processing. Based on these methods, HotSpotNet~\citep{chen2019object} and CenterPoint~\citep{yin2021center} introduce anchor-free approaches for better localizing 3D objects, and SA-SSD~\citep{he2020structure} proposes to use point-level supervisory signals to learn more discriminative voxel-based features. To improve the detection performance, Voxel R-CNN~\citep{deng2021voxel} first introduces a voxel-based two-stage approach using a voxel RoI pooling method. Focals Conv~\citep{chen2022focal} proposes focal sparse 3D convolutions to learn sparse voxel features by focusing more on important voxels at training time. Based on the success of vision transformer~(ViT), many ViT-based 3D object detectors have recently been introduced~\citep{he2022voxel, fan2022embracing,mao2021voxel,zhou2022centerformer}, showing better performance compared to CNN-based methods. The recent work of~\citep{he2024scatterformer} exploits linear attention to segmented voxel grids and introduces cross-window interaction with axis-decomposed convolutions to improve the voxel-based transformers.

Voxel-based 3D detection methods provide more precise 3D representations than BEV-based approaches, and provide better detection results in terms of accuracy. However, processing voxel grids and 3D convolutions in voxel-based methods is computationally demanding. In contrast to these methods, our method extracts voxel-based features efficiently, without using 3D convolutions. Voxel R-CNN~\citep{deng2021voxel} is closely related to ours in that it is two-stage voxel-based method that refines 3D proposals using voxel-based features directly. Differently, we produce final predictions using multi-level features from a backbone network and global contexts of a scene, making it possible to leverage contextual information better. In addition, our approach to using 2D CNNs is more efficient than Voxel R-CNN in terms of runtime and the number of parameters. 

\subsection{Point-based approach}
Point-based approaches~\citep{qi2018frustum,shi2019pointrcnn,wang2019frustum,yang2019std,yang20203dssd,shi2020pv,shi2020point,chen2019fast} learn 3D representations from point-based features obtained using PointNet-like networks~\citep{qi2017pointnet,qi2017pointnet++}. Adopting a two-stage approach, PointRCNN~\citep{shi2019pointrcnn} exploits raw point clouds directly to generate 3D proposals, and then refines the proposals using the point-based features inside them. Several approaches~\citep{yang2019std,shi2020pv,miao2021pvgnet,mao2021pyramid,sheng2021improving,chen2019fast} have followed the two-stage pipeline that introduce a new proposal generation method~\citep{yang2019std,miao2021pvgnet,chen2019fast} or an RoI refinement method~\citep{shi2020pv,sheng2021improving,mao2021pyramid,shi2023pv}. In contrast to this, other approaches~\citep{shi2020point,yang20203dssd} focus on single-stage 3D object detectors using point-based features. They propose a feature-based sampling strategy~\citep{yang20203dssd} or a graph-based local aggregation method~\citep{shi2020point} to generate reliable candidate points for predicting object bounding boxes. With the remarkable success of vision transformer~(ViT), a number of ViT-based 3D object detectors have recently been proposed~\citep{pan20213d,kolodiazhnyi2024unidet3d,wang2025state}. They introduce a transformer backbone to model both local and global relations~\citep{pan20213d}, a transformer decoder that enriches scene features~\citep{wang2025state}, or a transformer-based detector for multi-dataset learning with a unified label space across datasets~\citep{kolodiazhnyi2024unidet3d}.

Although point-based approaches preserve accurate 3D structures, they mainly have two problems, limiting the effectiveness and efficiency of 3D detectors, especially when handling a large-scale point cloud data. First, aggregating local information using a grouping method~\citep{qi2017pointnet++}, which is essential for feature extraction, is computationally expensive. Second, downsampling input point clouds in point-based approaches, due to the memory limitation, degrades the detection performance. To overcome these limitations, state-of-the-art methods~\citep{shi2020pv,miao2021pvgnet,mao2021pyramid,sheng2021improving} adopt a hybrid strategy exploiting voxel-based or BEV-based representations, along with the point-based features. For example, PV-RCNN and PV-RCNN++~\citep{shi2020pv, shi2023pv} introduce a VSA module to combine multi-scale features similar to S$^{2}$CFM in our approach. However, they do not consider global scene contexts for refining RoI features, and  require point-based operations and lots of network parameters to tune the module, which is computationally expensive.

\begin{figure*}[!t]
	% \captionsetup{font={small}}
	   \centering
		  \includegraphics[width=0.95\linewidth]{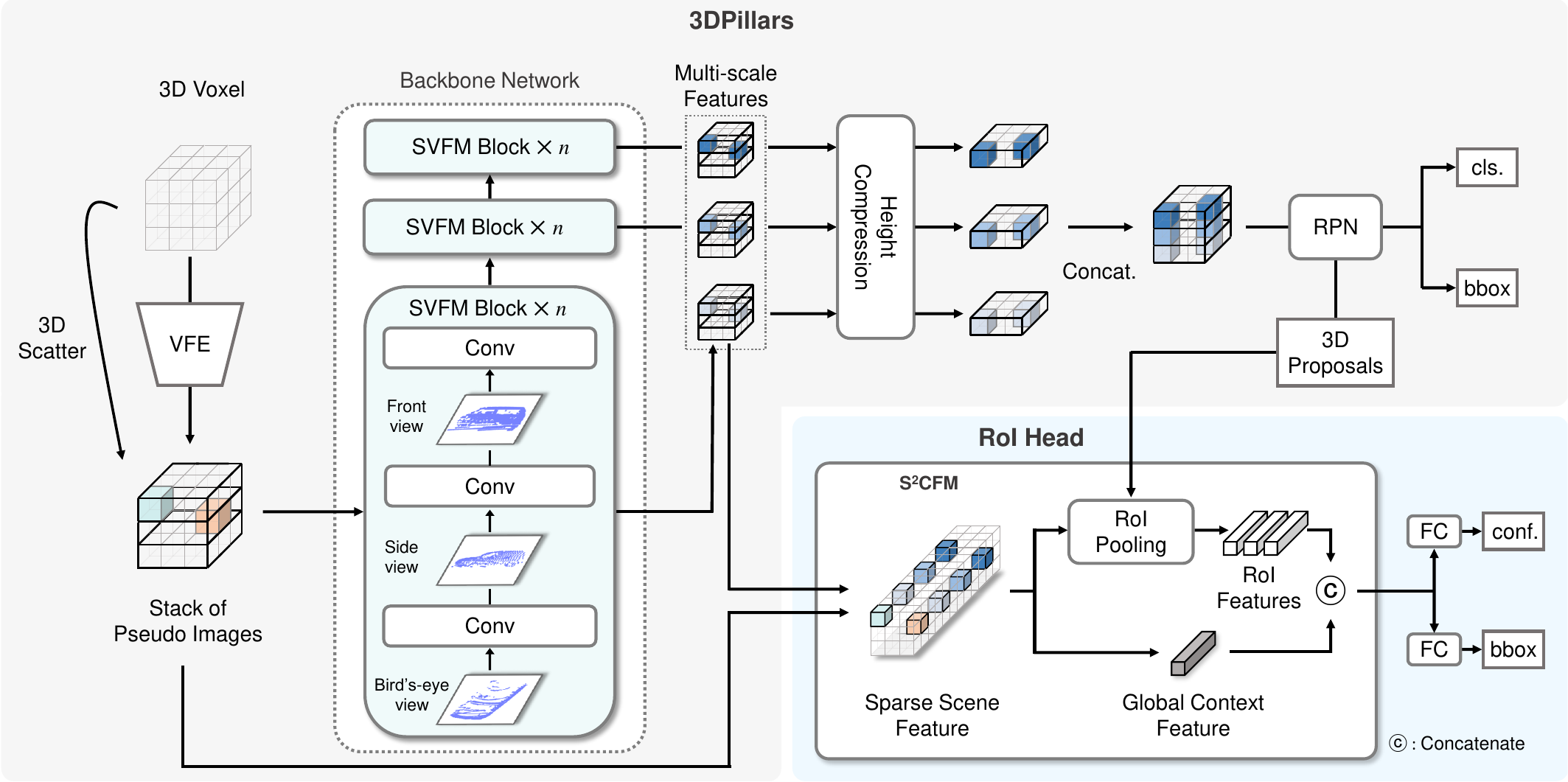}
  \caption{An overview of our approach. Our two-stage framework mainly consists of 3DPillars and an RoI head with S$^2$CFM. 3DPillars extracts multi-scale features from 3D voxels, without using 3D convolutions. To this end, SVFM considers the features from VFE as a stack of pseudo images, and extracts view-specific features by applying a 2D convolution to each slice of the stack along X-, Y- and Z-axes~(Fig.~\ref{fig:svfm}). 3DPillars further exploits multi-scale features from different levels of SVFM blocks, and generates 3D object proposals. S$^2$CFM in the RoI head then refines the proposals using sparse scene and global context features (Fig.~\ref{fig:roihead}). See text for details.}
   \label{fig:overview}
   \vspace{-0.4cm}
\end{figure*}

\section{Approach}

\noindent Our model mainly consists of two components~(Fig.~\ref{fig:overview}): 3DPillars~(Sec.~\ref{sec:3dpillar}), and an RoI head with S$^{2}$CFM~(Sec.~\ref{sec:roihead}). With input point clouds, 3DPillars outputs stacked view-specific representations that encode features from 3D voxels. To this end, we exploit a voxel feature encoding~(VFE) to extract 3D features from the voxels. The backbone network, consisting of a series of SVFM blocks, then extracts multi-scale features without using 3D convolutions. Specifically, considering the 3D features from VFE as a stack of pseudo images, SVFM splits them along each axis, and applies a 2D convolution to each slide separately to extract view-specific features~(Fig.~\ref{fig:svfm}). The multi-scale features are converted to 2D BEV feature maps, where a region proposal network~(RPN) generates 3D object proposals. The RoI head refines the proposals for final predictions. Specifically, S$^{2}$CFM combines the multi-scale features to obtain a sparse scene feature. It then pools RoI features, and refines them using global context features to predict 3D object bounding boxes and object confidences. In the following, we describe our two-stage detection framework in detail.
 
 \vspace{-0.1cm}

  \subsection{3DPillars}\label{sec:3dpillar}
  \vspace{0.1cm}
  \subsubsection{Voxel to multiple pseudo images} 
  Let us suppose that the size of each voxel is $v_{W} \times v_{L} \times v_{H}$. We define a 3D voxel grid of size $W/v_{W} \times L/v_{L} \times H/v_{H}$, where $W$, $L$, and $H$ are width, length, and height of a 3D space in a given scene, respectively. To generate 3D voxels, we assign input point clouds to the voxel grid according to 3D coordinates. Following~\citep{lang2019pointpillars}, we augment each point, together with a corresponding absolute coordinate and reflectance, using x-y-z offsets of the points and an average point for each voxel, w.r.t the center of the voxel. We then exploit VFE layers~\citep{zhou2018voxelnet}, a small version of PointNet~\citep{qi2017pointnet}, as our encoder to extract 3D features of size $D \times N$ from the augmented point clouds, where $D$ and $N$ are a dimension for each voxel and the number of non-empty voxels, respectively. We scatter the 3D features back to the corresponding locations in the 3D voxel grid to obtain a sparse feature volume of size $L/v_{L} \times W/v_{W} \times  H/v_{H} \times D$. By splitting the feature volume along each axis, we can obtain multiple view-dependent pseudo images, \eg, BEV pseudo images of size $L/v_{L} \times W/v_{W} \times D$. This enables maintaining fine-grained 3D structures using a stack of 2D representations for 3D object detection, and extracting 3D voxel features with a 2D convolutional backbone network, which will be discussed in the following.  

  \begin{figure}[!t]
	% \captionsetup{font={small}}
	   \centering
		  \includegraphics[width=1.0\linewidth]{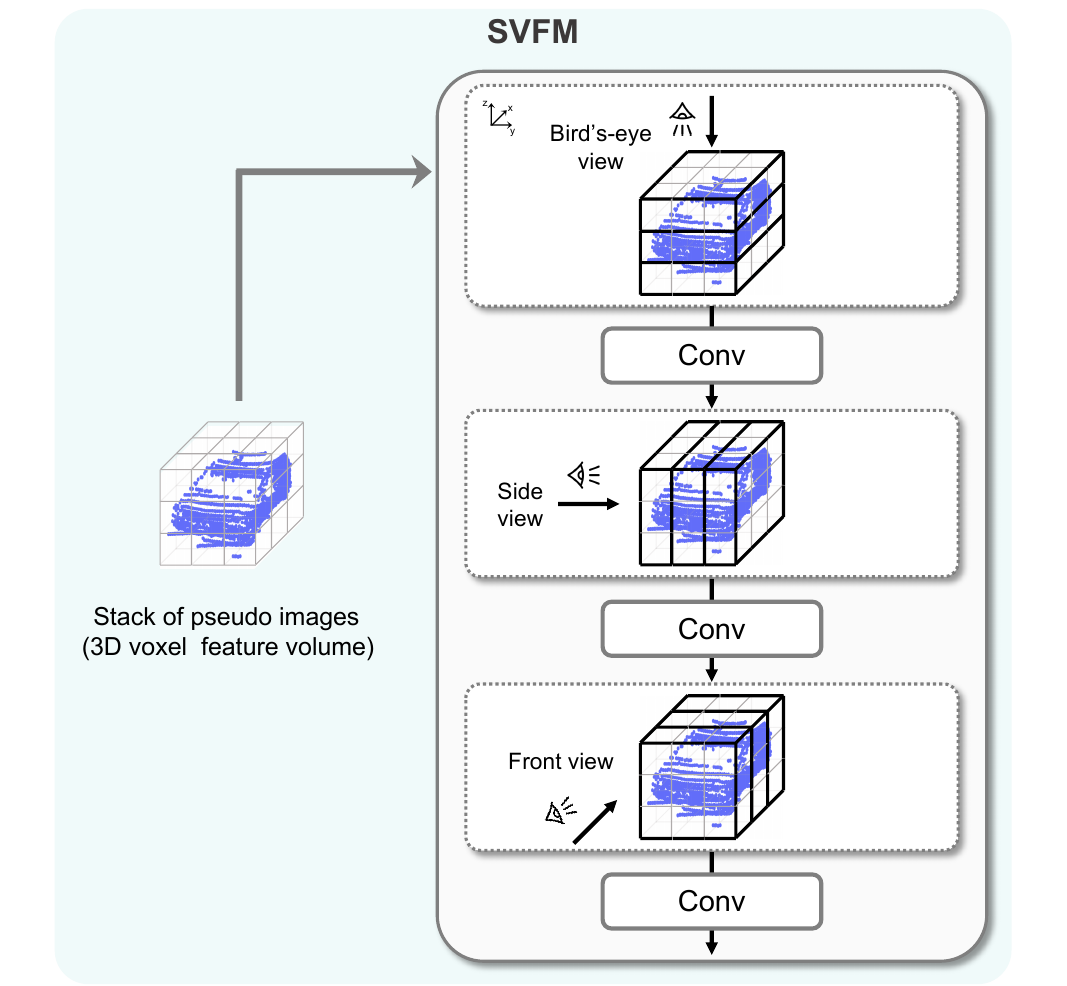}
  \caption{Illustration of SVFM. The sparse feature volume,~\ie,~scattered 3D features from VFE, can be represented as a stack of pseudo images. Based on this, SVFM decomposes the sparse feature volume along X-, Y-, and Z-axes to obtain stacks of pseudo images for front, side, and bird's-eye views, respectively, and applies 2D convolutions to each slide to extract view-specific features, allowing to learn voxel-based features without using 3D convolutions. See text for details.}
   \label{fig:svfm}
   \vspace{-0.4cm}
	\end{figure}

\vspace{-0.1cm}
  \subsubsection{Backbone network with SVFMs} 
    Our backbone network extracts voxel features encoding 3D structures efficiently, without using 3D convolutions as in other voxel-based approaches. To this end, SVFM splits the sparse feature volume,~\ie, a stack of pseudo images, along X-, Y-, and Z-axes, and applies 2D convolutions to each slide to extract view-specific features efficiently~(Fig.~\ref{fig:svfm}). For instance, to aggregate BEV-specific spatial information, we first reshape the 4D features from VFE to 3D spatial ones by collapsing the height dimension~(along the Z-axis), and apply 2D convolutions along the X-Y axes. Other view-specific information can be abstracted similarly. More specifically, instead of using 3D convolutional filters of size $k \times k \times k$, SVFM exploits three 2D convolutions, whose filter sizes are $1 \times k \times k$, $k \times 1 \times k$, and $k \times k \times 1$, respectively, to obtain BEV, side-view, and front-view specific features. We build a SVFM block by cascading SVFMs, which is a basic component of our backbone network~(Fig.~\ref{fig:overview}). Each SVFM block produces a backbone feature at certain scale. For the first SVFM in each block, we apply a convolution with a stride of 2 to gradually downsample the features. We use three SVFM blocks in the backbone, resulting in three voxel features with different scales. We squeeze the multi-scale features along the Z-axis to obtain 2D BEV feature maps, and concatenate them to generate 3D object proposals. 
    
    SVFM allows our backbone network to preserve 3D structures of a scene, while extracting voxel-based features without using 3D convolutions. It also enables learning view-to-view relationships effectively and efficiently by integrating the complementary information from each view within a 2D CNN framework. Obviously, exploiting 2D convolutions requires a much smaller number of parameters than 3D convolutions~(\ie,~a 45\% parameter reduction compared to the 3D case when $k=3$), which is computationally efficient.        
	
	\begin{figure}[!t]
		\centering
		\subfigure[Sequential]{\includegraphics[width=.23\textwidth]{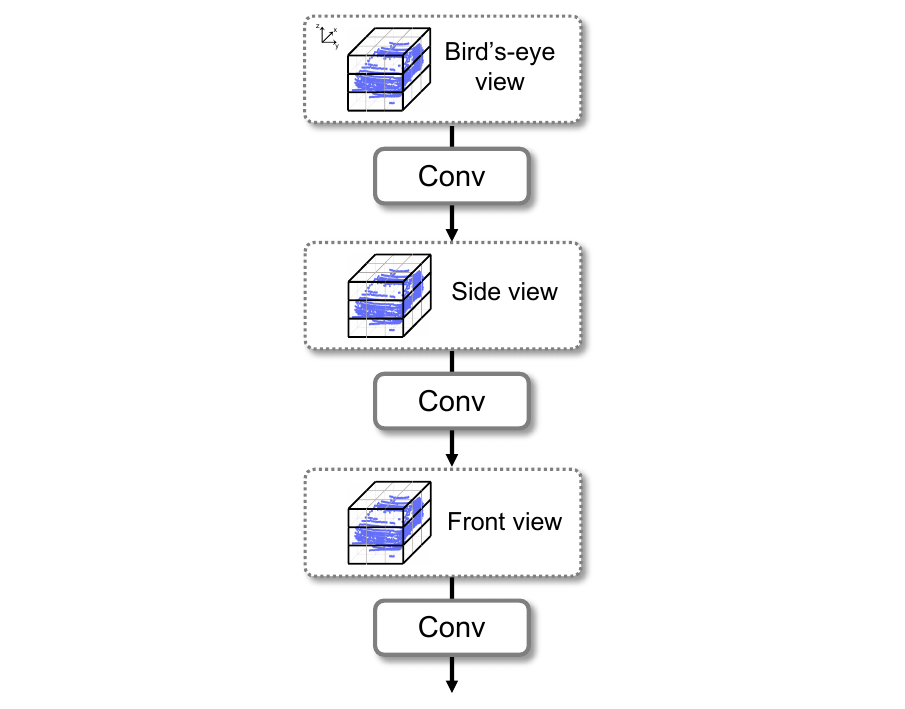}}
		\subfigure[Parallel]{\includegraphics[width=.23\textwidth]{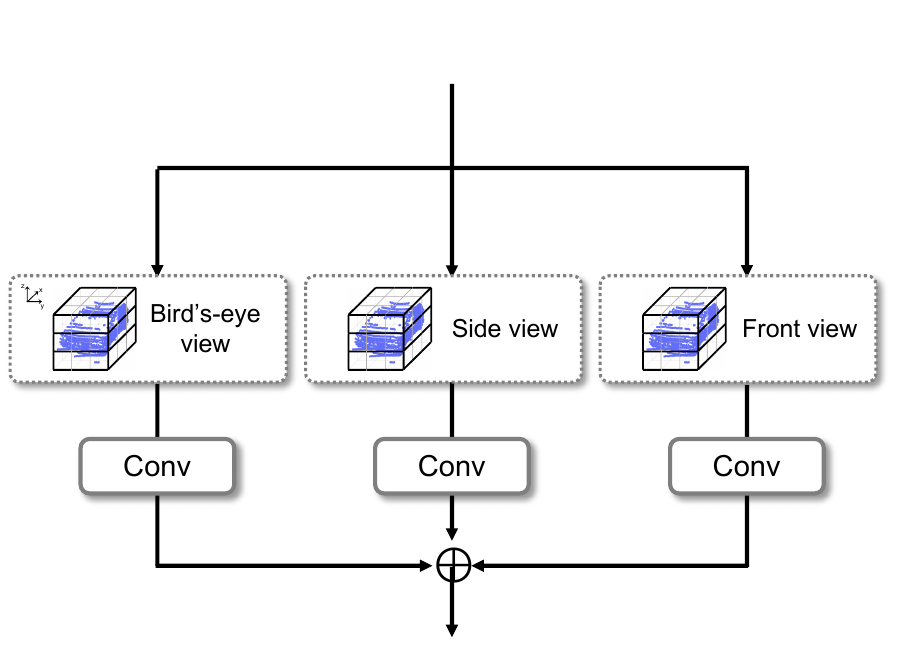}}\\
		\subfigure[Sequential-parallel]{\includegraphics[width=.23\textwidth]{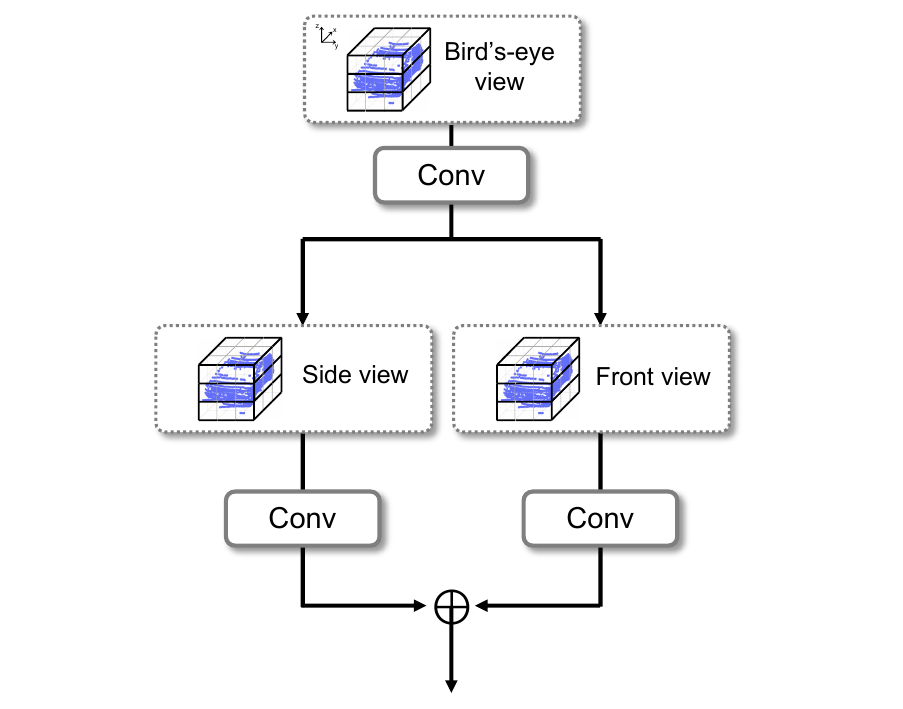}}
		\subfigure[Parallel-sequential]{\includegraphics[width=.23\textwidth]{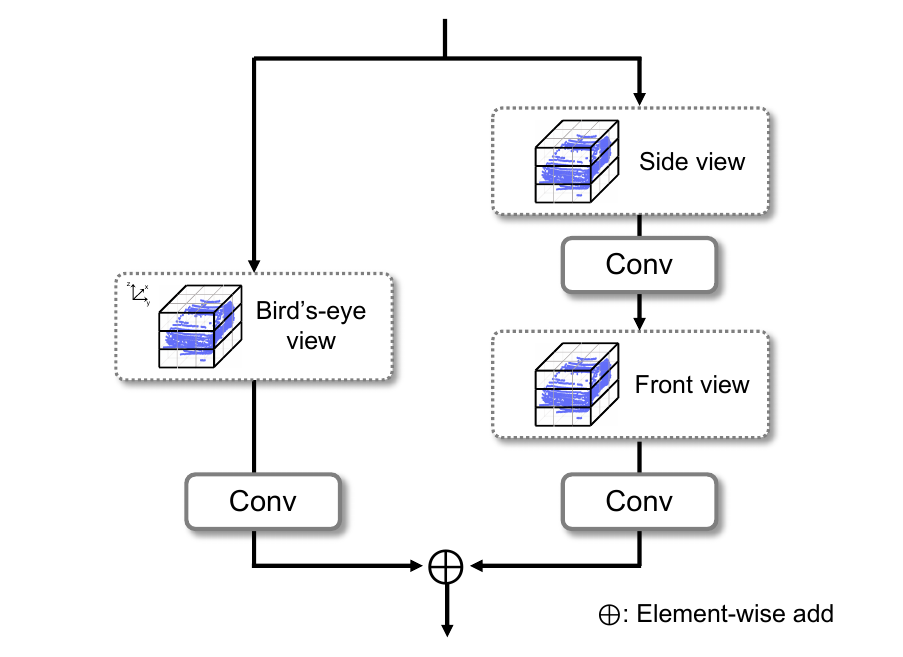}}
		% \vspace{-0.2cm}
	
		\caption{Illustration of different versions of SVFM. The sequential SVFM in~(a) learns view-specific features consecutively, while the parallel one in~(b) aggregates the features simultaneously. The sequential-parallel and parallel-sequential SVFMs in~(c) and~(d), respectively, consist of two parts that extract BEV-specific and height-related features. The former extracts the features sequentially, while the latter performs the feature extraction in parallel. See text for details.}
		\label{fig:svfm_blocks}
		\vspace{-0.4cm}
	
	  \end{figure}	

  \subsubsection{SVFM} 
  We introduce four versions of SVFM which differ in the order of extracting view-specific features~(Fig.\ref{fig:svfm_blocks}). Sequential and parallel SVFMs, shown in Fig.~\ref{fig:svfm_blocks}(a) and (b), respectively, extract features from view-dependent pseudo images either sequentially or simultaneously. The sequential SVFM is more effective to consider relationships between different views, while the parallel one offers better results in terms of efficiency. Other variants of SVFM, sequential-parallel and parallel-sequential SVFMs in Fig.~\ref{fig:svfm_blocks}(c) and (d), compromise between the sequential and parallel SVFMs. They process BEV-specific features and height-related features from side and front views separately. Specifically, the variants consist of two parts abstracting each complementary information individually, and connect the parts in series or in parallel similar to the sequential and parallel SVFMs. For our experiments, we mainly use the sequential SVFM~(Fig.\ref{fig:svfm_blocks}(a)), which is best suited for exploiting complementary information across views. This sequential structure promotes continuity and progressive context aggregation within each view. Although the views are processed independently, they share overlapping spatial dimensions (e.g., the Y-axis is common to both the bird’s-eye and side views), enabling the model to implicitly learn inter-view dependencies. As a result, both intra-view and inter-view spatial contexts are effectively captured. Empirically, we also observe that the sequential modules shown in Fig.~\ref{fig:svfm_blocks}(a) and (c) consistently outperform their parallel counterparts. See Sec.~\ref{sec:discussion} for more details.  
 
%  \begin{figure}[!t]
%   \centering
%   \subfloat[Sequential]{\includegraphics[width=.24\textwidth]{figures/svfm_1.pdf}}
%   \subfloat[Parallel]{\includegraphics[width=.24\textwidth]{figures/svfm_3.pdf}}\\
%   \subfloat[Sequential-parallel]{\includegraphics[width=.24\textwidth]{figures/svfm_2.pdf}}
%   \subfloat[Parallel-sequential]{\includegraphics[width=.24\textwidth]{figures/svfm_4.pdf}}
  
%   \caption{Illustration of different versions of SVFM. The sequential SVFM in~(a) learns view-specific features consecutively, while the parallel one in~(b) aggregates the features simultaneously. The sequential-parallel and parallel-sequential SVFMs in~(c) and~(d), respectively, consist of two parts that extract BEV-specific and height-related features. The former extracts the features sequentially, while the latter performs the feature extraction in parallel. See text for details.}
%   \label{fig:svfm_blocks}
% \end{figure}

  \begin{figure*}[t]
    % \captionsetup{font={footnotesize}}
       \centering
          \includegraphics[width=1.0\linewidth]{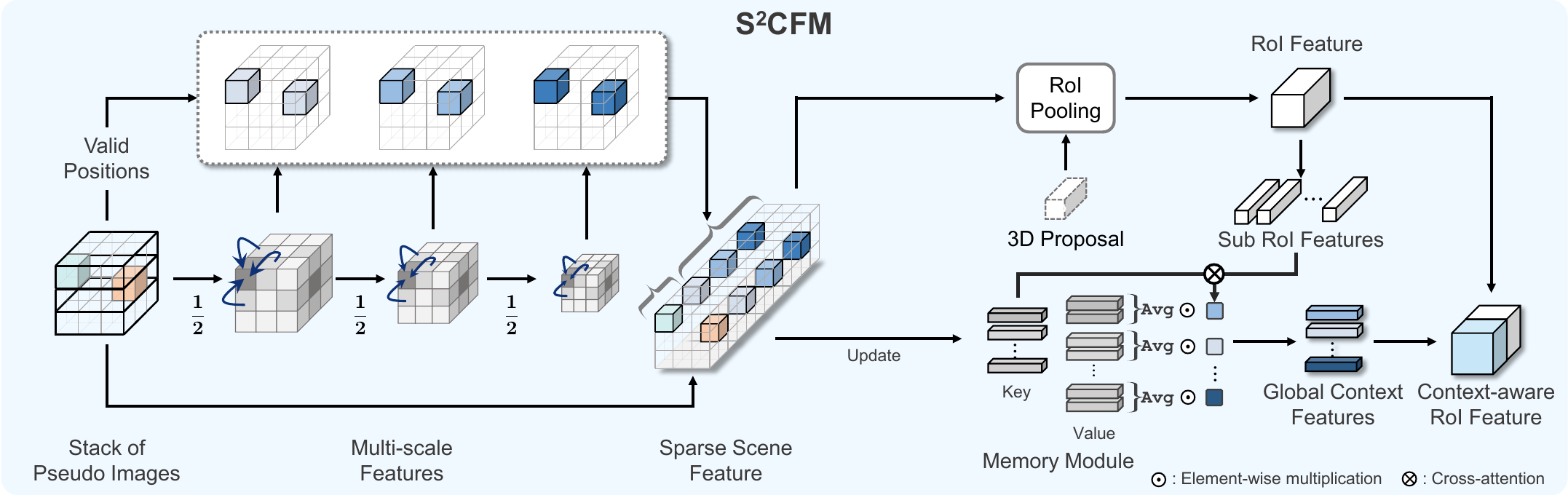}
  \caption{Illustration of S$^{2}$CFM. S$^{2}$CFM aggregates multi-scale features and a stack of pseudo images into a sparse scene feature. It then pools RoI features from the sparse scene feature and combines each RoI feature with global context features to obtain a context-aware RoI representation. To obtain the global context features, we augment the sparse scene feature with a key-value memory module, where each item stores the global contextual features of a scene.}
   \label{fig:roihead}
   \vspace{-0.5cm}
 \end{figure*}

  \subsection{RoI head with S$^{2}$CFM}\label{sec:roihead}
  Two-stage 3D object detectors refine 3D proposals using 3D RoI features to predict object bounding boxes, providing state-of-the-art detection results~\citep{deng2021voxel,mao2021pyramid,sheng2021improving,shi2023pv,hu2022density}. However, they exploit the features pooled from particular levels of a backbone only~\citep{deng2021voxel}. This might ignore useful features,~\eg,~for small objects typically obtained from shallow layers with large feature resolution~\citep{li2020netnet,kong2017ron,zhang2018single,noh2021hvpr}, which is problematic especially when predicting small objects. This refining strategy is also not generally applicable to BEV-based methods, since the 3D object proposals cannot be aligned with BEV features from a 2D convolutional backbone network. Exploiting rotated 3D boxes can be an alternative to pool BEV-based features for 3D RoI regions, but this cannot preserve precise 3D structures of the RoIs~\citep{deng2021voxel}. On the contrary, our model provides 3D voxel features with 2D convolutions, suggesting that the voxel RoI pooling method~\citep{deng2021voxel} might be applied for the second stage. However, the pooling method, designed for a sparse CNN, would not easily be integrated into our framework. In addition, it does not fully consider the global contextual information learned from the network to refine the proposals, as described in Sec.~\ref{sec:intro}. To address these problems, we introduce S$^{2}$CFM and plug it in the RoI head, allowing our BEV-based framework to benefit from the advantages of a two-stage pipeline. Our RoI head consists of S$^{2}$CFM module and two fully connected layers~(FC layers) for box regression and confidence prediction (Fig.~\ref{fig:overview}). S$^{2}$CFM is designed to enhance RoI features by incorporating both local geometric and global contextual information, and performs the following three operations (Fig.~\ref{fig:roihead}): 
  (1) It first aggregates multi-scale features extracted from different SVFM blocks, along with the initial voxel features from VFE, to construct a sparse yet contextually rich scene feature  which we refer to as a sparse scene feature.  
  (2) Next, for each 3D object proposal, a RoI feature is extracted from the sparse scene feature using the voxel RoI pooling method~\citep{deng2021voxel}, effectively capturing local geometric information.
  (3) Finally, the locally pooled RoI features are subdivided into smaller parts (Sub-RoI features), each of which queries a key-value memory module that stores prototypical global context features shared across scenes. Through a cross-attention mechanism, the memory retrieves context features relevant to the current RoI, which are then fused with the sub-RoI features to produce globally-informed, context-aware RoI representations.
  The resulting enriched RoI features are then passed through two FC layers to generate the final bounding box predictions and confidence scores. In the following, we describe each step in detail. 

  \subsubsection{RoI pooling}
  We obtain a sparse scene feature by concatenating the initial features from VFE~(\ie, a stack of pseudo images) and the multi-scale features from the backbone. To this end, we exploit the locations of non-empty voxels in the pseudo images. To be specific, we project 3D coordinates of the valid positions to the multi-scale features, and estimate features corresponding to the given coordinates by trilinear interpolation. We then concatenate the interpolated and initial features along the channel dimension to obtain the sparse scene feature of size $C \times N$, where RoI features are pooled. The sparse scene feature contains both the precise 3D structures in the initial features and rich contexts learned from the backbone network, which provides more discriminative RoI features that boost the detection performance significantly. We denote by $\mathbf{f}_\mathrm{scene}(n)$ each scene feature of size $C \times 1$, where $n=1,\cdots,N$. We use a voxel RoI pooling method in~\citep{deng2021voxel} to extract the RoI features. It divides each 3D proposal into $T \times T \times T$ sub-regions, and aggregates neighboring scene features into each sub-box, providing sub-RoI features of size $C \times T^{3}$. We denote by $\mathbf{f}_\mathrm{roi}(t)$ each sub-RoI feature of size $C \times 1$, where $t=1,\cdots,T^{3}$. 
   
  % \todo{describe the reason why we are using the locations for the concatenation}

  \subsubsection{Context-aware RoI feature}
    Previous two-stage methods refine 3D object proposals using the RoI features alone, leveraging local contextual information near the proposals only. This might not be sufficient to encode a global scene context for object detection, which is crucial for scene-level recognition tasks~\citep{zhang2018context,zhang2019co,qin2019thundernet,zhao2017pyramid,guo2020augfpn}. To address this problem, we augment each RoI feature with global contexts to produce context-aware RoI features. To our knowledge, it is the first attempt to consider the global scene context explicitly to refine the 3D proposals for 3D object detection.

 The global contexts can usually be represented as a single feature vector, obtained by~\eg,~global pooling, which is however not enough to model contextual information in complex outdoor scenes~\citep{geiger2012we,sun2020scalability}. In addition, similar environments~(\eg, road, sidewalk) commonly appear across the outdoor scenes. Based on these observations, we propose to augment the sparse scene feature with a memory module, shared across the all scenes, where each memory item stores various and universal prototypes of scene-level global contexts, and obtain global context features. Concretely, we define our memory module as a key-value memory~\citep{miller2016key} with key-addressing and value-reading schemes, which allows us to encode the diverse prototypical contexts of a scene more flexibly and efficiently than a single-type memory~\citep{weston2015memory,sukhbaatar2015end,gong2019memorizing}. The key memory is designed as a matrix $M_{k} \in \mathbb{R}^{K \times C}$ to store the $K$ representative prototypical contexts shared across all the scenes, where $K$ is the number of items in the key memory. For each key item, we define the value memory with a matrix $M_{v} \in \mathbb{R}^{V \times C}$, where $V$ is the number of items in each value memory. It encodes the diverse concepts of each prototype in the key memory. Using key and value memories, we obtain the context-aware RoI features as follows:

 \paragraph{Key-addressing}
 We first retrieve representative global contexts in the key memory with the RoI features. To this end, we compute matching probabilities between the key items and the sub-RoI features, as follows:
 \begin{equation}\label{eq:wk}
  {W}(t,k) = \frac{\exp(\mathbf{f}_\mathrm{roi}(t)^\top{\mathbf{f}}_\mathrm{key}(k))}{\sum_{k^\prime} \exp(\mathbf{f}_\mathrm{roi}(t)^\top{\mathbf{f}}_\mathrm{key}(k^\prime))},
  \vspace{-0.1cm}
\end{equation} 
where  $t=1,\cdots,T^{3}$, and $k=1,\cdots,K$. We denote by $\mathbf{f}_\mathrm{key}(k)$ each key item of size $C \times 1$. As shown in Eq.~\eqref{eq:wk}, the memory module retrieves the key items according to the similarities between the items and the sub-RoI features to augment the RoI features. This enables the network to simultaneously consider diverse global contextual features in the memory module.   

\paragraph{Value-reading} 
 For each key item, we aggregate corresponding value items to generate a mean value feature as follows:
 \begin{equation}\label{eq:wv}
  \mathbf{g}_\mathrm{value}(k) = \frac{1}{V} \sum_v \mathbf{f}_\mathrm{value}(k,v),    
  \vspace{-0.2cm}  
\end{equation} 
%Although there may be other options for aggregation, such as weighted summation~\citep{miller2016key}, we equally associate the value items to generate the global context features. This imposes the same weights for the value items, allowing us to consider diverse prototypes better.
where $k=1,\cdots,K$, and $v=1,\cdots,V$. We denote by $\mathbf{f}_\mathrm{value}(k,v)$ and $\mathbf{g}_\mathrm{value}(k)$ each value item of size $C \times 1$ and its average of size $C \times 1$, respectively. We then integrate the features $\mathbf{g}_\mathrm{value}(k)$ using the addressing probabilities $W(t,k)$ for each sub-RoI feature: 
  \begin{equation}\label{eq:wg}
     \mathbf{g}_\mathrm{roi}(t) = \sum_k {W}(t,k) \mathbf{g}_\mathrm{value}(k),      
     \vspace{-0.2cm}
  \end{equation}
  where $\mathbf{g}_\mathrm{roi}(t)$ is an aggregated global context feature.
  Finally, we obtain the context-aware RoI feature by concatenating the aggregated context features and the sub-RoI ones along the channel dimension, and input the RoI feature to the FC layers to predict 3D bounding boxes and confidence values for corresponding classes. This imposes richer contexts from various but similar scenes to the RoI features, while providing diverse global contexts for complex scenes. Our RoI head is thus able to refine the 3D object proposals considering diverse global contexts, particularly effective for the objects with sparse point clouds,~\ie, small in size or distant. 

\subsubsection{Memory update}
  We leverage the sparse scene feature to encode diverse and prototypical contexts in the memory module. To update items in the key memory, we first compute matching probabilities between the key items and the scene features as in Eq.~\eqref{eq:wk}, we then select all scene features showing the highest probability for each item. Note that multiple scene features can be assigned to a single item in the key memory. We then encourage an average of the selected features and memory items to be similar as follows:  
  \begin{equation}\label{eq:keyloss}
    \mathcal{L}_\mathrm{key} = \sum_k \|\mathbf{f}_\mathrm{key}(k)-\mathbf{\hat{f}}_\mathrm{scene}(k)\|_2,
    \vspace{-0.2cm}
 \end{equation}
where $\Vert \cdot \Vert_{2}$ computes the L2 norm of a vector, and $\mathbf{\hat{f}}_\mathrm{scene}$ is an average of the the selected scene features, defined as follows:
\begin{equation}\label{eq:keytarget}
  \mathbf{\hat{f}}_\mathrm{scene}(k) = \frac{1}{L} \sum_{l\in U_k} \mathbf{f}_\mathrm{scene}(l).
  \vspace{-0.2cm}
\end{equation} 
We denote by $U_k$ the set of indices for the corresponding scene features for the $k$-th item in the key memory, and $L$ is the number of the corresponding features. 
Furthermore, to promote diversity of the prototypes, we encourage each key item to have unique prototypes as follows:
\begin{equation}\label{eq:ortholoss}
  \mathcal{L}_\mathrm{ortho} = \|{I}_{K}-M_{k}M_{k}^T\|_2,
  \vspace{-0.2cm}
\end{equation}
where we denote by ${I}_K$ an identity matrix of size~$K\times K$.
These two losses make the key items to restore the most representative prototypes of the contexts, shared by all the scenes, since the outdoor scenes can share some common scene-level global contexts.  
To update the value memory, we exploit a subset of the selected scene features~($\mathbf{f}_\mathrm{scene}(l), l\in U_k$). We first sort the selected scene features in a descending order based on the matching probabilities in Eq.~\eqref{eq:wk}. We then subsample the features of the top-$V$ highest probabilities. Finally, we encourage the items of the value memory and the sampled scene features to be similar as follows:
\vspace{-0.1cm}
\begin{equation}\label{eq:valueloss}
  \mathcal{L}_\mathrm{value} = \sum_k\sum_{v} \|\mathbf{f}_\mathrm{value}(k,v)-\mathbf{\bar{f}}_\mathrm{scene}(k,v)\|_2, 
  \vspace{-0.2cm}
\end{equation}
where $v=1 \dots V$ and we denote by $\mathbf{\bar{f}}_\mathrm{scene}$ the scene features sorted in a descending order. This enables each item in the value memory to record diverse representations of the prototypes in the key memory. To sum up, we update our memory module as follows:
\vspace{-0.2cm}
\begin{equation}\label{eq:memloss}
  \mathcal{L}_\mathrm{mem} = \mathcal{L}_\mathrm{key} + \mathcal{L}_\mathrm{ortho} + \mathcal{L}_\mathrm{value}.
  \vspace{-0.2cm}
\end{equation}

  % This helps the RoI head produce better predictions by leveraging learning about the scene by increasing the receptive field.

  % To alleviate these issues, S$^{2}$CFM aggregates all multi-scale backbone features into a single sparse scene feature volume, where the RoI features are pooled, by leveraging the locations of valid voxels. This allows rich scene context to be effectively and efficiently encoded in the RoI features, which boosts the 3D detection performance significantly. Given the RoI features, S$^{2}$CFM further refines them using global context features of the 3D scene, which are particularly effective for localizing small objects.

  \subsection{Training loss}

  Following~\citep{shi2020pv,deng2021voxel}, we apply RPN and RoI losses to the outputs of the RPN and the RoI head, respectively. First, the RPN loss consists of regression~($L_{\mathrm{reg}}$), angle classification~($L_{\mathrm{dir}}$), and classification~($L_{\mathrm{cls}}$) terms as follows:
  \begin{equation} \label{loss:RPN}
    \mathcal{L}_{\mathrm{RPN}}=\frac{1}{N_{p}} \left( \lambda_{\mathrm{reg}} \mathcal{L}_{\mathrm{reg}} + \lambda_{\mathrm{dir}} \mathcal{L}_{\mathrm{dir}} + \lambda_{\mathrm{cls}} \mathcal{L}_{\mathrm{cls}} \right),
    \vspace{-0.2cm}
  \end{equation}
  where $N_{p}$ is the number of positive anchors and $\lambda_{\mathrm{reg}}$, $\lambda_{\mathrm{dir}}$, and $\lambda_{\mathrm{cls}}$ are balancing parameters for each term. The regression term~($\mathcal{L}_{\mathrm{reg}}$) applies a smooth L1 function to residual parameters between anchor and ground-truth boxes in the RPN, where the residual parameters are defined with the center position, the size, and the heading angle of a box. The angle classification term~($\mathcal{L}_{dir}$) prevents penalizing the same 3D boxes with opposite directions~\citep{yan2018second}. We adopt the focal loss~\citep{lin2017focal} for the classification term~($\mathcal{L}_{cls}$). Second, the RoI loss consists of regression~($\tilde{\mathcal{L}}_{\mathrm{reg}}$), confidence~($\mathcal{L}_{\mathrm{cfd}}$), and memory~($\mathcal{L}_{\mathrm{mem}}$) terms as follows:
  \begin{equation} \label{loss:RoI}
    \mathcal{L}_{\mathrm{RoI}}=\frac{1}{N_{s}} \left( \mathbb{I} [\text{IoU}\ge \gamma] \tilde{\mathcal{L}}_{\mathrm{reg}} + \mathcal{L}_{\mathrm{cfd}} + \lambda_{\mathrm{mem}} \mathcal{L}_{\mathrm{mem}} \right),
    \vspace{-0.2cm}
  \end{equation}
  where~$N_s$ is the number of region proposals in the RoI head. $\mathbb{I} [\text{IoU}\ge \gamma]$ outputs one if a region proposal shows a IoU score greater than or equal to the threshold $\gamma$ and zero otherwise. The regression term~($\tilde{\mathcal{L}}_{\mathrm{reg}}$) in Eq.~\eqref{loss:RoI} is defined analogous to the one in Eq.~\eqref{loss:RPN}, but it exploits predictions from the RoI head. The confidence term~($\mathcal{L}_{\mathrm{cfd}}$) is a binary cross-entropy loss for a confidence prediction, where the target confidence is computed using IoU thresholds on foreground and background regions~\citep{deng2021voxel}.

\vspace{-0.2cm}
   \section{Experiments}
   In this section, we describe implementation details, and show experimental comparisons of our model and the state of the art. We mainly exploit a sequential SVFM module for our model~(Fig.~\ref{fig:svfm_blocks}(a)), unless otherwise specified.
   
	 \subsection{Implementation details}
	 \vspace{0.2cm}
	 \subsubsection{Datasets and evaluation protocols}
	 We evaluate our model on the standard 3D object detection benchmarks, the KITTI dataset~\citep{geiger2012we}, the nuScenes dataset~\citep{caesar2020nuscenes}, and the Waymo Open dataset~\citep{sun2020scalability}. KITTI provides 7,481 and 7,518 samples for training and test, respectively. Following~\citep{noh2021hvpr,liu2020tanet}, we divide the training samples into training and validation splits with a ratio of around 5:1, to evaluate our model on an official evaluation server. Otherwise, we split the training samples into 3,712 and 3,769 samples for training and validation, following the work of~\citep{chen20153d}. Following a standard evaluation protocol, we report the mean average precision~(mAP) with 40 recall positions on a PR curve. For comparison on the validation set, the mAP is calculated with 11 recall positions for a fair comparison with other methods. The nuScenes dataset~\citep{caesar2020nuscenes} consists of 1,000 driving sequences, with 700, 150, and 150 sequences for training, validation, and testing, respectively. It employs a 32-beam LiDAR sensor, producing around 30K points per frame. Following the official evaluation protocol, we report mean Average Precision~(mAP) and nuScenes Detection Score~(NDS). To densify the point clouds, we accumulate 10 consecutive LiDAR sweeps. The Waymo Open dataset contains 798 and 202 sequences for training and validation, respectively, with around 158K and 40K point clouds. We use mAP and mAP weighted by heading accuracy~(mAPH) as evaluation metrics for Waymo Open. To measure the performance, we use all validation sequences and official evaluation tools. The testing samples are divided according to difficulty levels~(\ie, LEVEL\_1 for boxes with more than five LiDAR points and LEVEL\_2 for boxes with at least one LiDAR point).  We apply non maximum suppression~(NMS) with a threshold of 0.1 to keep top-100 proposals for both datasets.

	\subsubsection{Training details}
	We train our network for 100 epochs, with a learning rate of 1e-2 and a batch size of 16 for the KITTI dataset, and train our network for 20 epochs, with a learning rate of 1e-3 and a batch size of 16 for the nuScenes dataset. For the Waymo Open dataset, the network is trained for 50 epochs, with a learning rate of 3e-3 and a batch size of 16. We use the one cycle learning policy~\citep{smith2019super} as a learning rate scheduler. We adopt augmentation techniques for point clouds, including random flipping along the X-axis, random scaling, and random rotation along the Z-axis. A scaling factor and a rotation angle are randomly chosen within a range of $[0.95, 1.05]$ and $[-\frac{\pi}{4}, \frac{\pi}{4}]$, respectively. We also use an augmentation technique for the ground truth that pastes randomly sampled ground-truth bounding boxes from other scenes~\citep{yan2018second}. 
   %  All models are trained end-to-end using \texttt{PyTorch}~\citep{paszke2017automatic}.  
   
	 \subsubsection{Parameter settings}
   Following PointPillars~\citep{lang2019pointpillars}, the dimension of a 3D scene~($W,H,L$) is within a range of $[(0,70.4), (-40,40),(-3,1)]$ meters and the voxel size of X-Y dimensions ($v_W,v_H$) is set to $(0.16, 0.16)$ for the KITTI dataset~\citep{geiger2012we}. For the Z dimension, we set the voxel size~$v_L$ as $0.25$. For the nuScenes dataset, the dimension of a scene is within a range of $[ (-54, 54), (-54, 54), (-5,3) ]$ meters. We set the voxel size of X-Y dimensions ($v_W,v_H$) to $( 0.2 , 0.2 )$, and the voxel size of $v_L$ to $0.4$. On the Waymo Open dataset, we follow previous pillar-based methods~\citep{shi2022pillarnet, lang2019pointpillars} by assuming that the dimension is within a range of $[(-75.2,75.2), (-75.2,75.2),(-2,4)]$ meters, and the voxel size of X-Y dimensions ($v_W,v_H$) is set as $(0.16, 0.16)$. We set the voxel size~$v_L$ as $0.3$ for the Z dimension. For a fair comparison, we use the same settings as in~\citep{deng2021voxel,shi2020pv,shi2023pv,lang2019pointpillars} for other parameters, including the number of voxels~($N$), the number of sub-RoI regions~($T$), and balancing parameters~($\lambda_{\mathrm{reg}}$, $\lambda_{\mathrm{dir}}$, and $\lambda_{\mathrm{cls}}$). The balancing parameter for the memory loss~($\lambda_\mathrm{mem}$) is set to~$0.5$. 
      
   \setlength{\tabcolsep}{0.03em}
   \begin{table*}[!ht]
   % \captionsetup{font={small}}
   \begin{center}
   
   \caption{Quantitative comparison with the state of the art in terms of mAP(\%) and runtime on the test set of KITTI~\citep{geiger2012we}. We categorize 3D object detection methods into three classes according to the types of 3D representations. Numbers in bold indicate the best performance and underscored ones are the second best for each type of approaches.}
	%   \begin{adjustbox}{width=1.0\columnwidth,center}
		\begin{footnotesize}
		 \begin{tabular}{c c c C{3em} C{3em} C{3em} C{3em} C{3em} C{3em} C{3em}}
		 \hline
		 {\multirow{2}{*}{Type}}& \multirow{2}{*}{Stage}&\multirow{2}{*}{Methods} &\multirow{2}{*}{\shortstack{Speed\\(Hz)}}  & \multicolumn{3}{c}{Car} &\multicolumn{3}{c}{Cyclist}  \\
		 &&&  &Easy & Mod. & Hard & Easy & Mod. & Hard    \\
		 \hline\midrule
		 {\multirow{7}{*}{Point~}} 
		 % &Two&PointRCNN~\citep{shi2019pointrcnn} &10.0 &86.96  &75.64  &70.70 &73.93  &59.60  &53.59   \\
		 % &Two&FastPointRCNN~\citep{chen2019fast}&\underline{16.7}  &85.29  &77.40  &70.24 &-  &-  &-  \\
		 % &Two&STD~\citep{yang2019std} &12.5  &87.95  &79.71  &75.09 &78.69 &61.59  &55.30     \\
		 &One&3DSSD~\citep{yang20203dssd} &\textbf{25.0}  & 88.36 &79.57  &74.55 &\textbf{82.48}  &\underline{64.10}  &\underline{58.00} \\
		%  &Two&Part-A2~\citep{shi2020points} &11.8 &85.94 &77.86  &72.00 &78.58  &62.73  &57.74   \\
		 &Two&PV-RCNN~\citep{shi2020pv}& ~~8.9&\textbf{90.25}  &81.43  &76.82 &78.60  &63.71  &57.65     \\
		 &One&PointGNN~\citep{shi2020point}&~~1.6 &88.33  &79.47  &72.29 &78.60  &63.48  &57.08    \\
		 &Two&Pyramid-PV~\citep{mao2021pyramid}& ~~7.9 &88.39  &\textbf{82.08}  &\textbf{77.49} &-  &-  &-  \\
		 &Two&CT3D~\citep{sheng2021improving} &\underline{14.3} &87.83  &81.77  &\underline{77.16} &-  &-  &-     \\
		 &Two&PV-RCNN++~\citep{shi2023pv}& 13.6& \underline{90.14}  &\underline{81.88}  &77.15 &\underline{82.22}  &\textbf{67.33}  &\textbf{60.04}     \\
		 \midrule
		 {\multirow{9}{*}{Voxel~}}
		 % &One&VoxelNet~\citep{zhou2018voxelnet}& ~~4.5& 77.47 &65.11  &57.73 &61.22 &48.36  &44.37   \\
		 % &One&SECOND~\citep{yan2018second} & 20.0& 83.13 & 73.66 &66.20 &70.51  &53.85  &46.90     \\
		 % &One&3DIoULoss~\citep{zhou2019iou}&12.5 & 86.16 & 76.50  &71.39 &-  &-  &-   \\
		 % &One&SA-SSD~\citep{he2020structure} &25.0 &88.75  &79.79  & 74.16 &-  &-  &-   \\
		 &One&HotSpotNet~\citep{chen2019object} &25.0 &87.60 &78.31  &73.34 &79.09  &62.72  &56.76   \\
		 &Two&Voxel R-CNN~\citep{deng2021voxel} &\underline{25.2} &\textbf{90.90} &81.62  &77.06 &-  &-  &-   \\
		 &Two&VoTR-TSD~\citep{mao2021voxel} & ~~7.2 &89.90  &{82.09}  &\textbf{79.14} &-  &-  &-   \\
		 &Two&PDV~\citep{hu2022density} &~~7.4 &{90.43} &81.86  &77.36 &\textbf{83.04}  &\underline{67.81}  &\underline{60.46}   \\
		 &Two&Focals Conv~\citep{chen2022focal} &~~8.9 &90.20 &{82.12}  &\underline{77.50} &-  &-  &-   \\
		%  &Two&VoxSeT~\citep{he2022voxel} &18.0 &88.53 &82.06  &77.46 &-  &-  &-   \\
		 &One&RDIoU~\citep{sheng2022rethinking} &\textbf{26.6} &\underline{90.65} &\textbf{82.30}  &77.26 &-  &-  &-   \\
		 &Two&PG-RCNN~\citep{koo2023pg} &16.6 &89.38 &82.13 &77.33 &\underline{82.77}  &\textbf{67.82}  &\textbf{61.25}   \\
		 &One&PVT-SSD~\citep{yang2023pvt} &20.4 &\underline{90.65} &\underline{82.29} &76.85 &-  &-  &-   \\
		 \midrule
		 {\multirow{6}{*}{BEV~}}
		 &One&PointPillars~\citep{lang2019pointpillars}& \textbf{42.4}& 82.58 &74.31  &68.99 &77.10  &58.65  &51.92  \\
		 &One&TANet~\citep{liu2020tanet} & 28.5& 84.39& 75.94 & 68.82 &73.84  &59.86 &53.46    \\
		%  &One&Associate-3D~\citep{du2020associate} &20.0& 85.99 & 77.40 & 70.53 &-  &-  &-    \\
		 &One&HVPR~\citep{noh2021hvpr}&  \underline{36.1}  &86.38  & 77.92 & 73.04 &-  &-  &-  \\
		 &Two&LiDAR R-CNN~\citep{li2021lidar}&  35.6  &85.97  & 74.21 & 69.18 &-  &-  &-  \\
		 &One&PillarNet-18~\citep{shi2022pillarnet}&  34.5  &\textbf{89.65}  & \underline{81.06} & \underline{76.67} &\underline{83.01}  &\underline{67.33}  &\underline{61.07}\\ 
		 \cmidrule(rl){2-10}
		 % &Two&\todo{Ours-SEQ-PAR} & 27.2 &88.41  & 81.68 & 76.93 &\textbf{83.69}  &\textbf{67.86}  &{60.54} \\
		 &Two&{Ours} & 29.6  &\underline{89.08}  & \textbf{81.83 }& \textbf{77.15} &\textbf{83.46}  &\textbf{67.71}  &\textbf{61.41}   \\
		 \hline
		 \end{tabular}
		\end{footnotesize}
		% \end{adjustbox}
	  \label{table:kitti}
   \end{center}
   \vspace{-0.5cm}
   \end{table*}

   % \vspace{-0.2cm}
   \subsubsection{Network architecture} 
   For each voxel, we exploit a VFE layer~\citep{zhou2018voxelnet} that consists of FC and Batchnorm~\citep{ioffe2015batch} layers to obtain a voxel-based feature, whose dimension~($D$) is set to~$32$, following~\citep{zhou2018voxelnet}. Our backbone network consists of three SVFM blocks, where each block has two SVFMs except the last block with three SVFMs. Following PointPillars~\citep{lang2019pointpillars}, we set the dimensions of output features as 64, 128, and 256, respectively, for each block, and the filter size is set to $3$ for all layers. We convert the multi-scale features to 2D BEV features maps, and upsample them with a transposed convolution such that the BEV features have the same spatial resolution as the largest one~(Fig.~\ref{fig:overview}). To generate 3D object proposals, we use two FC layers with $576$ channels. We apply a $1 \times 1$ convolutional layer to reduce the channel dimension of the sparse scene feature~($C$) to $160$. To obtain RoI features, we first find neighboring scene features for each sub-RoI region by selecting up to 32 voxel features within two thresholds, determined by the Manhattan distance of 2 and 4 as in~\citep{deng2021voxel}. We then use two PointNet layers~\citep{qi2017pointnet} with channel sizes of 32 and 16, respectively, to aggregate those neighboring features. The RoI head consists of two FC layers and two branches for box regression and confidence prediction, respectively. The channel sizes are set to 256 for all layers in the detection head. We determine the sizes of key memory~($K$) and value memory~($V$) for the KITTI~\citep{geiger2012we} dataset by a grid search in terms of the mAP on the validation split, setting them to $10$ and~$50$, respectively. We use the same network parameters for the Waymo Open dataset~\citep{sun2020scalability}, except for the memory sizes~($K$ and $V$). Considering the number of training samples in the Waymo Open dataset, we adjust the sizes of the key and value memories to $20$ and $200$, respectively.      

   \setlength{\tabcolsep}{0.2em}
	 \begin{table*}[!ht]
	 % \captionsetup{font={small}}
	 \begin{center}
	 \caption{Quantitative comparison with the state of the art in terms of mAP(\%) and runtime on the validation set of KITTI~\citep{geiger2012we}. * denotes the reproduced results by ours. Numbers in bold indicate the best performance and underscored ones are the second best.}
		% \begin{adjustbox}{width=1.0\columnwidth,center}
		\vspace{-0.1cm}
		\begin{footnotesize} 
		   \begin{tabular}{c c c C{2.9em} C{2.9em} C{2.9em} C{2.9em} C{2.9em} C{2.9em} C{2.9em} C{2.9em} C{2.9em} C{2.9em}}
		   \hline
		   {\multirow{2}{*}{Type}}& \multirow{2}{*}{Stage}&\multirow{2}{*}{Methods} &\multirow{2}{*}{\shortstack{Speed\\(Hz)}}  & \multicolumn{3}{c}{Car} &\multicolumn{3}{c}{Cyclist} &\multicolumn{3}{c}{Pedestrian} \\
		   &&&  &Easy & Mod. & Hard & Easy & Mod. & Hard & Easy & Mod. & Hard    \\
		   \hline\midrule
		   {\multirow{7}{*}{Point~}} 
		   % &Two&PointRCNN~\citep{shi2019pointrcnn} &10.0 &88.88  &78.63  &77.38 &-  &72.11  &-  &-  &54.41  &-    \\
		   % &Two&FastPointRCNN~\citep{chen2019fast}&\underline{16.7}  &85.29  &77.40  &70.24 &-  &-  &- &-  &-  &-  \\
		   % &Two&STD~\citep{yang2019std} &12.5  &89.70  &79.80  &79.30 &\underline{88.50} &\underline{72.80}  &67.90  &\textbf{73.90} &\textbf{66.60}  &\textbf{62.90}    \\
		   &One&3DSSD~\citep{yang20203dssd} &\textbf{25.0}  & 88.36 &79.57  &74.55 &-  &-  &- &-  & - &-  \\
		   &Two&Part-A2~\citep{shi2020points} &11.8 &89.47 &79.47  &78.54 &\underline{88.31}  &\textbf{73.07}  &\textbf{70.20} &\textbf{70.37}  &\textbf{65.99}  &\textbf{57.48}   \\
		   &Two&PV-RCNN~\citep{shi2020pv}& ~~8.9& 89.33  &83.69  &78.72 &86.06  &70.47  &64.55 &64.10  &57.90  &52.77   \\
		%    &One&PointGNN~\citep{shi2020point}&~~1.6 &87.89  &78.34  &77.38 &-  &-  &- &-  &-  &-  \\
		   &Two&Pyramid-PV~\citep{mao2021pyramid}& ~~7.9 &\underline{89.37}  &84.38  &\underline{78.84} &-  &-  &-   &-  &-  &-  \\
		   &Two&CT3D~\citep{sheng2021improving} &\underline{14.3} &{89.11}  &\textbf{85.04}  &{78.76} &{85.04}  &71.71  &68.05  &- &-  &-    \\
		   &Two&PV-RCNN++*~\citep{shi2023pv}& 13.6&\textbf{89.57}  &\underline{84.83}  &\textbf{78.86} &\textbf{91.01}  &\underline{72.55}  &\underline{69.18} &\underline{64.23}  &\underline{58.26}  &\underline{53.78}    \\
		   \midrule
		   {\multirow{8}{*}{Voxel~}}
		   % &One&VoxelNet~\citep{zhou2018voxelnet}& ~~4.5& 81.97 &65.46  &62.85 &67.17 &47.65  &45.11   &57.86 &53.42  &48.87  \\
		   % &One&SECOND~\citep{yan2018second} & 20.0& 88.13 & 78.62 &73.20 &83.51  &67.15  &64.90  &55.51  &52.98  &46.90   \\
		   % &One&3DIoULoss~\citep{zhou2019iou}&12.5 & 89.15 & 79.14  &78.11 &-  &-  &-  &-  &-  &-    \\
		   % &One&SA-SSD~\citep{he2020structure} &25.0 &\textbf{90.15}  &79.91  & 78.78 &-  &-  &-   &-  &-  &-  \\
		   &One&HotSpotNet~\citep{chen2019object} &25.0 &87.60 &78.31  &73.34 &\underline{88.22}  &72.55  &68.08 &\textbf{72.23}  &\textbf{65.90}  &\textbf{60.06}   \\
		   &Two&Voxel R-CNN~\citep{deng2021voxel} &25.2 &89.41 &84.52  &\underline{78.93} &{85.98} &{72.56}  &{69.14} & \underline{68.29}  &\underline{61.74}  &\underline{59.24}   \\
		%    &Two&VoTR-TSD~\citep{mao2021voxel} & ~~7.2 &89.04  &84.04  &78.68 &-  &-  &- &-  &-  &-    \\
		   &Two&PDV*~\citep{hu2022density} &~~7.4 &\underline{89.43}  &84.10  &{78.89} &86.83  &\underline{73.07}  &\underline{71.03} &63.96  & 58.58 &53.87   \\
		   &Two&Focals Conv~\citep{chen2022focal} &~~8.9 &\textbf{89.52} &\underline{84.93}  &\textbf{79.18} &-  &-  &- &-  &-  &-   \\
		   &One&RDIoU~\citep{sheng2022rethinking} &\underline{26.6} &89.16 &\textbf{85.24}  &78.41 &83.32  &68.39  &63.63 &63.26  &57.47  &52.53   \\
		   &One&VoxSeT~\citep{he2022voxel} &\textbf{29.4} &88.45 &78.48  &77.07 &84.07  &68.11  &65.14 &60.62  &54.74  &50.39   \\
		   &Two&PG-RCNN*~\citep{koo2023pg} &16.6 &\textbf{89.52} &84.84  &78.70 &\textbf{92.32}  &\textbf{73.34}  &\textbf{71.40} &64.69  &59.46  &58.29   \\
		   \midrule
		   {\multirow{6}{*}{BEV~}}
		   &One&PointPillars~\citep{lang2019pointpillars}& \textbf{42.4}& 87.53 &76.88  &74.04 &80.91  &64.03  &\underline{60.75} &58.51  &54.13  &48.36 \\
		   &One&TANet~\citep{liu2020tanet} & 28.5& 88.21& 77.85 & 75.62 &{85.98}  &\underline{64.95} &{60.40}  &\underline{70.80}  &\underline{63.45} &\underline{58.22}    \\
		   &One&Associate-3D~\citep{du2020associate} &20.0& \underline{89.29} & 79.17 & \underline{77.76} &-  &-  &-  &-  &-  &-   \\
		   % &One&HVNet~\citep{ye2020hvnet} & 31.0 & 87.21 & 77.58 & 71.79 &-  &-  &- &-  &-  &-  \\
		   &One&HVPR~\citep{noh2021hvpr}&  \underline{36.1}  &88.54  & \underline{79.94} & 77.20 &\underline{86.06}  &64.55  &59.58  &63.43  &57.45  &52.82  \\
		   &One&PillarNet-18*~\citep{shi2022pillarnet}& 34.5  &88.04  & 78.65 & 75.45 &78.38  &63.56  &59.85  &63.25  &57.09  &52.76  \\ 
		   &One&{PillarNeXt-B*}~\citep{li2023pillarnext}& 33.3  &88.45 &78.90 & 76.07 &80.79  &63.36  &60.11  &64.30 &58.67  &56.40  \\ 
		   \cmidrule(rl){2-13}
	 %        &Two&Ours-SEQ-PAR & 28.5  &{89.51}  & 84.80 & 78.83 &91.79  &{73.11}  &{70.50} &71.48  &\underline{68.46}  &\underline{61.91}   \\
		   &Two&Ours & 29.6  &\textbf{89.63}  & \textbf{84.79} & \textbf{78.87} &\textbf{92.43}  &\textbf{73.56}  &\textbf{69.96} &\textbf{75.08}  &\textbf{68.97}  &\textbf{62.55}   \\
		   
		   \hline
		   \end{tabular}
		\end{footnotesize}
		% \end{adjustbox}
		\label{table:kittival}
		\vspace{-0.8cm}

	 \end{center}	 
	 \end{table*}
   
	 \subsection{Results}
	 \vspace{0.2cm}
	 \subsubsection{Quantitative comparisons on KITTI}
	 We compare in Table~\ref{table:kitti} our method with the state of the art for LiDAR-based 3D object detection on the test set of KITTI~\citep{geiger2012we}. We evaluate our method on car and cyclist classes, most widely used ones for comparisons. We report mAP scores with IoU thresholds of $0.7$ and $0.5$ for car and cyclist, respectively. All mAP scores for other methods are taken from corresponding papers or~an official evaluation server. We categorize 3D object detection methods into three groups according to the types of 3D representations (point-based, voxel-based, and BEV-based approaches), and list them in each type in chronological order of publications. We can observe three things from Table~\ref{table:kitti} as follows: (1) Our model boosts the detection performance of PointPillars~\citep{lang2019pointpillars} remarkably. For example, mAP gains of $6.50, 7.52, 8.16$ on easy, moderate, and hard difficulty levels for the car class, respectively. For the cyclist class, the gains are $6.36, 9.06, 9.49$ on the easy, moderate, and hard splits, respectively. Our model also outperforms other BEV-based methods in terms of all evaluation metrics with a large margin. This demonstrates the effectiveness of our two-stage framework that overcomes the limitations of the pillar-based methods.    
	 (2) Our method outperforms previous state-of-the-art methods for the cyclist class, while providing competitive results for the car class, indicating that it successfully narrows the performance gaps between the pillar-based method~\citep{lang2019pointpillars} and the state of the art. Although recent hybrid point-based methods~\citep{shi2023pv,mao2021pyramid,sheng2021improving}, two-stage voxel-based methods~\citep{hu2022density,chen2022focal}, and transformer-based methods~\citep{mao2021voxel,he2022voxel,yang2023pvt,sheng2022rethinking} achieve higher accuracy for the car class, our method is much faster than these approaches, offering a real-time inference speed of 29.6~Hz. This is particularly notable in the context of real-time applications, which typically require a minimum speed of 20~Hz~\citep{ye2020hvnet,simony2018complex,geiger2012we}, especially considering that the KITTI dataset provides annotations only within a 90-degree field of view. Among all methods, ours achieves the best mAP across all classes at real-time speed, while many state-of-the-art point-based and voxel-based detectors~\citep{shi2023pv,sheng2021improving,yang2023pvt,koo2023pg} operate below the real-time threshold of 20~Hz.
	 As will be shown in Sec~\ref{sec:discussion}, our model also has fewer parameters than most state-of-the-art approaches. This suggests that our method offers a good compromise in terms of speed and accuracy.
	 (3) The performance improvements over PointPillars~\citep{lang2019pointpillars} are particularly significant for the moderate and hard splits of the car class, where 3D objects are typically small in size or distant. Similarly, our method gives more improvements on the moderate and hard splits of the cyclist class, and the improvements for the cyclist class are much larger than the car class. This confirms that augmenting RoI features with global contexts helps to extract more discriminative features from the object proposals, especially composed of sparse point clouds. For further comparisons, we show in Table~\ref{table:kittival} our model with the state of the art on the KITTI validation set for three classes. We report mAP scores with IoU thresholds of 0.7, 0.5, and 0.5 for car, cyclist, and pedestrian, respectively, obtained with 11 recall positions. We can observe consistent findings with the results on the test set. We can see that our method improves the detection performance of PointPillars~\citep{lang2019pointpillars} significantly, outperforming other BEV-based methods. The improvements for the cyclist and pedestrian classes, small in size, are more significant than the ones for the car class. Our method also gives the best results on the cyclist and pedestrian classes, while the performances for the car class are on a par with the state-of-the-art methods.

	 \vspace{-0.2cm}

	 \setlength{\tabcolsep}{0.2em}
	 \begin{table*}[!t]
		 % \captionsetup{font={small}}
		 \begin{center}
		 \caption{Quantitative comparison with the state of the art in terms of 3D mAP(\%) and mAPH(\%) on the validation set of Waymo Open Dataset~\citep{sun2020scalability}. Numbers in bold indicate the best performance and underscored ones are the second best.}
		   \begin{adjustbox}{width=2.1\columnwidth,center}
			\vspace{-0.1cm}
		   \begin{footnotesize}
			% \begin{tabular}{c c c C{2.9em} C{2.9em} C{2.9em} C{2.9em} C{2.9em} C{2.9em} C{2.9em} C{2.9em} C{2.9em} C{2.9em}}
			   \begin{tabular}{c c c c c c c c c c c c}
			   \hline
			   \multirow{2}{*}{Type}& \multirow{2}{*}{Stage}&\multirow{2}{*}{Methods} &\multirow{2}{*}{\shortstack{Vehicle~(L1)\\mAP/mAPH}} &\multirow{2}{*}{\shortstack{Vehicle~(L2)\\mAP/mAPH}}& \multirow{2}{*}{\shortstack{Cyclist~(L1)\\mAP/mAPH}} & \multirow{2}{*}{\shortstack{Cyclist~(L2)\\mAP/mAPH}} & \multirow{2}{*}{\shortstack{Pedestrian~(L1)\\mAP/mAPH}} & \multirow{2}{*}{\shortstack{Pedestrian~(L2)\\mAP/mAPH}}\\
			   &&&  & &  &  &  & \\
			   \hline\midrule
			   {\multirow{5}{*}{Point~}} 
			   &Two&Part-A2~\citep{shi2020points}&71.69/71.16 &64.21/63.70 & 68.60/67.36 &66.13/64.93 & 75.24/66.87 &66.18/58.62  \\
			   &Two&PV-RCNN~\citep{shi2020pv}&\underline{78.00}/\underline{77.50} &\underline{69.43}/\underline{68.98}  &\underline{71.46}/\underline{70.27} &\underline{68.95}/\underline{67.79}  &\underline{79.21}/\underline{73.03}  &\underline{70.42}/\underline{64.72}     \\
			%    &Two&Pyramid-PV~\citep{mao2021pyramid}&{76.30}/{75.68} &{67.23}/{66.68}  &\hspace{0.3cm}-\hspace{0.3cm}~/\hspace{0.3cm}-\hspace{0.3cm}~ &\hspace{0.3cm}-\hspace{0.3cm}~/\hspace{0.3cm}-\hspace{0.3cm}~ &\hspace{0.3cm}-\hspace{0.3cm}~/\hspace{0.3cm}-\hspace{0.3cm}~ &\hspace{0.3cm}-\hspace{0.3cm}~/\hspace{0.3cm}-\hspace{0.3cm}~     \\
			%    &Two&CT3D~\citep{sheng2021improving}&\hspace{0.16cm}{76.30}/\hspace{0.3cm}-\hspace{0.3cm}~ &\hspace{0.16cm}{69.04}/\hspace{0.3cm}-\hspace{0.3cm}~  &\hspace{0.3cm}-\hspace{0.3cm}~/\hspace{0.3cm}-\hspace{0.3cm}~ &\hspace{0.3cm}-\hspace{0.3cm}~/\hspace{0.3cm}-\hspace{0.3cm}~ &\hspace{0.3cm}-\hspace{0.3cm}~/\hspace{0.3cm}-\hspace{0.3cm}~ &\hspace{0.3cm}-\hspace{0.3cm}~/\hspace{0.3cm}-\hspace{0.3cm}~   \\
			   &Two&PV-RCNN++~\citep{shi2023pv}&\textbf{79.25}/\textbf{78.78} &\textbf{70.61}/\textbf{70.18}  &\textbf{73.72}/\textbf{72.66} &\textbf{71.21}/\textbf{70.19}  &\textbf{81.83}/\textbf{76.28}  &\textbf{73.17}/\textbf{68.00}  \\
			   \midrule
			   {\multirow{10}{*}{Voxel~}}
			   &One&SECOND~\citep{yan2018second}&72.27/71.69 &63.85/63.33 & 60.62/59.28 &58.34/57.05 & 68.70/58.18 &60.72/51.31  \\
			   &Two&CenterPoint~\citep{yin2021center}&76.70/76.20 &68.80/68.30  &\hspace{0.3cm}-\hspace{0.3cm}~/\hspace{0.3cm}-\hspace{0.3cm}~ &\hspace{0.3cm}-\hspace{0.3cm}~/68.6 &{79.00}/{72.90} &71.00/65.30  \\
			   &Two&Voxel R-CNN~\citep{deng2021voxel} &{76.13}/75.66 &68.18/67.74 &{70.75}/{69.68}  &{68.25}/{67.21} &78.20/71.98 &69.29/63.59  \\
			%    &Two&VoTR-SSD~\citep{mao2021voxel}&68.99/68.39 &60.22/59.69  &\hspace{0.3cm}-\hspace{0.3cm}~/\hspace{0.3cm}-\hspace{0.3cm}~ &\hspace{0.3cm}-\hspace{0.3cm}~/\hspace{0.3cm}-\hspace{0.3cm}~ &\hspace{0.3cm}-\hspace{0.3cm}~/\hspace{0.3cm}-\hspace{0.3cm}~ &\hspace{0.3cm}-\hspace{0.3cm}~/\hspace{0.3cm}-\hspace{0.3cm}~    \\
			
			   &Two&CenterFormer~\citep{zhou2022centerformer}&75.00/74.40 &69.90/{69.40}  &{73.80}/{72.70} &{71.30}/{70.20} & 78.00/72.40 &{73.10}/{67.70}  \\ 
			   &Two&SST\textunderscore TS\textunderscore 1f~\citep{fan2022embracing}&76.22/75.79 &68.04/67.64  &\hspace{0.3cm}-\hspace{0.3cm}/\hspace{0.3cm}-\hspace{0.3cm}~&\hspace{0.3cm}-\hspace{0.3cm}/\hspace{0.3cm}-\hspace{0.3cm}~ & {81.39}/{74.05} &{72.82}/{65.93} \\
			   &Two&PDV~\citep{hu2022density}&76.85/{76.33} &69.30/68.61  &68.71/67.55 &66.49/65.36  &74.19/65.96  &65.85/58.28  \\
			 &Two&VoxSeT+CT3D~\citep{he2022voxel}&\hspace{0.16cm}{77.82}/\hspace{0.3cm}-\hspace{0.3cm}~ &\hspace{0.16cm}{70.21}/\hspace{0.3cm}-\hspace{0.3cm}~  &\hspace{0.3cm}-\hspace{0.3cm}~/\hspace{0.3cm}-\hspace{0.3cm}~ &\hspace{0.3cm}-\hspace{0.3cm}~/\hspace{0.3cm}-\hspace{0.3cm}~ &\hspace{0.3cm}-\hspace{0.3cm}~/\hspace{0.3cm}-\hspace{0.3cm}~ &\hspace{0.3cm}-\hspace{0.3cm}~/\hspace{0.3cm}-\hspace{0.3cm}~  \\
			   &One&PVT-SSD~\citep{yang2023pvt}&{79.16}/{78.72} &{70.23}/{69.83}  &\hspace{0.3cm}-\hspace{0.3cm}~/\hspace{0.3cm}-\hspace{0.3cm}~ &{73.94}/{72.96}&\hspace{0.3cm}-\hspace{0.3cm}~/\hspace{0.3cm}-\hspace{0.3cm}~ &72.56/67.02 \\
			      &One&ScatterFormer~\citep{he2024scatterformer}&\textbf{81.00}/\textbf{80.50} & \textbf{73.10}/\textbf{72.70} & \underline{79.90}/\underline{78.90} & \underline{77.10}/\underline{76.10} & \underline{84.50}/\underline{79.90} & \underline{77.00}/\underline{72.60} \\
				  &Two&SAFDNet~\citep{zhang2024safdnet}&\underline{80.60}/\underline{80.10} &\underline{72.70}/\underline{72.30}  &\textbf{80.00}/\textbf{79.00} & 
				  \textbf{77.20}/\textbf{76.20} & 
				  \textbf{84.70}/\textbf{80.40} & 
				  \textbf{77.30}/\textbf{73.10}  \\
			   \midrule
			   {\multirow{7}{*}{BEV~}}
			   &One&PointPillars(RSN impl.)~\citep{lang2019pointpillars}&63.30/62.70 & 55.20/54.70&\hspace{0.3cm}-\hspace{0.3cm}/\hspace{0.3cm}-\hspace{0.3cm}~&\hspace{0.3cm}-\hspace{0.3cm}/\hspace{0.3cm}-\hspace{0.3cm}~& 68.90/56.60 &60.00/49.10\\
			   &One&Pillar-OD~\citep{wang2020pillar} & \hspace{0.16cm}69.80~/\hspace{0.3cm}-\hspace{0.3cm}~&\hspace{0.3cm}-\hspace{0.3cm}/\hspace{0.3cm}-\hspace{0.3cm}~& \hspace{0.3cm}-\hspace{0.3cm}/\hspace{0.3cm}-\hspace{0.3cm}~&\hspace{0.3cm}-\hspace{0.3cm}/\hspace{0.3cm}-\hspace{0.3cm}~& \hspace{0.16cm}72.51/\hspace{0.3cm}-\hspace{0.3cm}~ & \hspace{0.3cm}-\hspace{0.3cm}/\hspace{0.3cm}-\hspace{0.3cm}~  \\
			   &Two&LiDAR R-CNN~\citep{li2021lidar}&73.50/73.00 &64.70/64.20  &68.60/66.90 & 66.10/64.40 &71.20/58.70  &63.10/51.70     \\
			   &One&PillarNet-18~\citep{shi2022pillarnet}&78.24/77.73 &\underline{70.40}/\underline{69.92}  &70.40/69.29&67.75/66.68& 79.80/72.59 & 71.57/64.90   \\
			   &One&SWFormer\textunderscore 1f~\citep{sun2022swformer}&77.80/77.30 &69.20/68.80  &\hspace{0.3cm}-\hspace{0.3cm}/\hspace{0.3cm}-\hspace{0.3cm}~&\hspace{0.3cm}-\hspace{0.3cm}/\hspace{0.3cm}-\hspace{0.3cm}~& 80.90/72.70 & 72.50/64.90  \\
			   &Two&CenterPoint-S2D~\citep{wang2022sparsedense}&76.10/75.53 & 68.11/67.58  &67.81/66.22&65.28/63.74& 74.29/65.20 & 66.41/58.06    \\
			   &One&PillarNeXt-B~\citep{li2023pillarnext}&\underline{78.40}/\underline{77.90} & 70.27/69.81  &\underline{73.21}/\underline{72.20}&\underline{70.58}/\underline{69.62}& \textbf{82.53}/\textbf{77.14} & \textbf{74.90}/\textbf{69.80}    \\
			   \cmidrule(rl){2-9}
			   &Two&{Ours} & \textbf{79.10}/\textbf{78.65}  &\textbf{70.42}/\textbf{70.00}  & \textbf{73.52}/\textbf{72.37} & \textbf{70.93}/\textbf{69.83} &\underline{81.58}/\underline{75.27} &\underline{73.12}/\underline{67.22}  \\
			   % &Two&\changed{Ours-SEQ-PAR} & 75.54/\underline{74.75}  &66.61/\underline{65.89}  & 92.41/\underline{91.82} & 73.99/\underline{73.11} &52.67/\underline{51.29} &34.30/25.70  \\
			   \hline
			 \end{tabular}
			\end{footnotesize}
		   \end{adjustbox}
		   \label{table:waymo}
		 \end{center}	
		   \vspace{-0.6cm}
	 \end{table*}

	 \setlength{\tabcolsep}{0.2em}
	 \begin{table*}[!ht]
	  %   \captionsetup{font={small}}
		\begin{center}
		\caption{Quantitative comparison with the state of the art in terms of 3D mAP(\%) and mAPH(\%) on the validation set of Waymo Open Dataset~\citep{sun2020scalability} for the vehicle class. Numbers in bold indicate the best performance and underscored ones are the second best.}
		\vspace{0.1cm}
		%    \begin{adjustbox}{width=1\columnwidth,center}
		\begin{footnotesize}
			  \begin{tabular}{c c c  C{6.5em} C{6.5em} C{6.5em}}
			  \hline
			  \multirow{2}{*}{Type}& \multirow{2}{*}{Stage}&\multirow{2}{*}{Methods} &\multicolumn{3}{c}{LEVEL\_1 mAP/mAPH by Distance} \\
			  &&&  0-30m & 30-50m & 50m-Inf \\
			  \hline\midrule
			  {\multirow{4}{*}{Point~}} 
			 %  &One&StarNet~\citep{ngiam2019starnet} &83.30/82.40  &58.80/53.20 &34.30/25.70     \\
			 &Two&PV-RCNN~\citep{shi2020pv}  &91.83/\underline{91.37}  &69.99/\underline{69.37} &46.26/\underline{45.41}     \\
			  &Two&Pyramid-PV~\citep{mao2021pyramid}&\underline{92.67}/\textbf{92.20}  &{74.91}/\textbf{74.21} &{54.54}/\textbf{53.45}     \\
			  &Two&CT3D~\citep{sheng2021improving} &{92.51}/\hspace{0.3cm}-\hspace{0.3cm}~ &\underline{75.07}/\hspace{0.3cm}-\hspace{0.3cm}~ &\underline{55.36}/\hspace{0.3cm}-\hspace{0.3cm}~    \\
			  &Two&PV-RCNN++~\citep{shi2023pv}&\textbf{93.05}/\hspace{0.3cm}-\hspace{0.3cm}~  &\textbf{77.70}/\hspace{0.3cm}-\hspace{0.3cm}~ &\textbf{57.38}/\hspace{0.3cm}-\hspace{0.3cm}~ \\
			  \midrule
			  {\multirow{5}{*}{Voxel~}}
			  &Two&Part-A2~\citep{shi2020points}& 91.83/91.37 &69.99/69.37 &46.26/45.41  \\
			 %  &One&CVCNet~\citep{chen2020every}  &86.80/\hspace{0.3cm}-\hspace{0.3cm}~  & 62.19/\hspace{0.3cm}-\hspace{0.3cm}~ &29.27/\hspace{0.3cm}-\hspace{0.3cm}~  \\
			  &Two&Voxel R-CNN~\citep{deng2021voxel}  &\underline{92.49}/\hspace{0.3cm}-\hspace{0.3cm}~  &\underline{74.09}/\hspace{0.3cm}-\hspace{0.3cm}~ &\underline{53.15}/\hspace{0.3cm}-\hspace{0.3cm}~  \\
			  &Two&VoTR-SSD~\citep{mao2021voxel}&88.18/87.62  &66.73/66.05&42.08/41.38  \\
			  &Two&VoTR-TSD~\citep{mao2021voxel} &92.28/\underline{91.73}  &73.36/\underline{72.56} &51.09/\underline{50.01}     \\
			  &Two&PDV~\citep{hu2022density}&\textbf{93.13}/\textbf{92.71}  &\textbf{75.49}/\textbf{74.91}&\textbf{54.75}/\textbf{53.90} \\
			  \midrule
			  {\multirow{3}{*}{BEV~}}
			  &One&PointPillars(RSN impl.)~\citep{lang2019pointpillars} &84.90/84.40  &59.20/58.60 &35.80/35.20 \\
			  &One&Pillar-OD~\citep{wang2020pillar} &88.53/\hspace{0.3cm}-\hspace{0.3cm}~ & 66.50/\hspace{0.3cm}-\hspace{0.3cm}~ &42.93/\hspace{0.3cm}-\hspace{0.3cm}~  \\
			  &Two&LiDAR R-CNN~\citep{li2021lidar}&\underline{92.10}/\underline{91.60}  &\underline{74.60}/\underline{74.10} &\underline{54.50}/\underline{53.40}     \\
			 %  &One&\todo{PillarNeXt-B}*~\citep{li2023pillarnext}&-/- &-/- &-/- \\
			  \cmidrule(rl){2-6}
			  &Two&{Ours} &\textbf{93.30}/\textbf{92.92} & \textbf{78.11}/\textbf{77.61} &\textbf{57.54}/\textbf{56.86}  \\
			  \hline
			  \end{tabular}
			\end{footnotesize}
		%    \end{adjustbox}
		   \label{table:waymo_car}
		\end{center}	
		   \vspace{-0.6cm}
	 \end{table*}
   
   % \vspace{-0.1cm}
   \subsubsection{Quantitative comparisons on Waymo Open and nuScenes}
	 To further validate the effectiveness of our approach, we compare in Table~\ref{table:waymo} our method with the state of the art on the large-scale Waymo Open dataset. We evaluate our model on the vehicle, cyclist and pedestrian classes, and report mAP and mAPH scores with an IoU threshold of $0.7$, $0.5$ and $0.5$, respectively. From this table, we can observe similar findings in Table~\ref{table:kitti}. Our method brings significant improvements over PointPillars~\citep{lang2019pointpillars} for all metrics, outperforming other pillar-based methods. Ours also gives comparable results with the state-of-the-art methods~\citep{shi2023pv, yang2023pvt}, while providing better results than the hybrid point-based approaches~\citep{shi2020pv,mao2021pyramid, sheng2021improving} and transformer-based voxel approaches~\citep{zhou2022centerformer, he2022voxel, fan2022embracing, mao2021voxel}. However, the most recent voxel-based methods~\citep{he2024scatterformer,zhang2024safdnet}, which fully leverage the sparse nature of point cloud inputs, achieve significantly better performance. Nonetheless, we believe that incorporating sparsity-aware mechanisms into our 2D convolutional framework could further close the gap with recent voxel-based detectors, while retaining the computational benefits of our approach. We also present in Table~\ref{table:waymo_car} the distance-based detection results for the vehicle class. The perfomance gains over PoinPillars~\citep{lang2019pointpillars} are particularly significant for the objects at farther distances,~\ie, 50m-Inf, which suggests that S$^{2}$CFM effectively captures rich contexts of a scene from the large-scale dataset.
	 The experimental results on the Waymo Open Dataset show that our approach reduces performance gaps between pillar-based detectors and high-performing ones effectively on the large-scale dataset, validating the generalization ability of our approach. 

	 To further evaluate the generalizability of our method beyond the KITTI and Waymo datasets, we conduct additional experiments on the nuScenes dataset~\citep{caesar2020nuscenes}. We mainly compare our method with previous BEV-based methods~\citep{lang2019pointpillars, yin2021center, shi2022pillarnet, li2023pillarnext}. As shown in Table~\ref{tab:nuscenes}, our method achieves the best performance in both 3D mAP (63.43\%) and NDS (68.96\%), outperforming all other BEV-based detectors. In terms of True Positive (TP) metrics, our method achieves either the best or highly competitive results across the board. These improvements demonstrate that our two-stage framework not only generalizes well to diverse and challenging datasets but also effectively addresses the limitations of previous BEV-based methods.

	 \setlength{\tabcolsep}{0.3em}
	 \begin{table*}[!t]
	 \centering
	 \caption{Quantitative comparison with the state of the art BEV methods in terms of 3D mAP(\%) and NDS(\%) on the validation set of nuScenes~\citep{caesar2020nuscenes}. Numbers in bold indicate the best performance and underscored ones are the second best.}
	 \label{tab:nuscenes}
	 \begin{footnotesize}
	 \begin{tabular}{c c l c c c c c c c c}
	 \hline
	 {Type} & {Stage} & {Method} & {mATE$\downarrow$} & {mASE$\downarrow$} & {mAOE$\downarrow$} & {mAVE$\downarrow$} & {mAAE$\downarrow$} & {mAP$\uparrow$} & {NDS$\uparrow$} \\
	 \hline \midrule
	 \multirow{5}{*}{BEV} 
	 & One & PointPillar-MultiHead~\citep{lang2019pointpillars}     & 33.87 & 26.00 & 32.07 & 28.74 & 20.15 & 44.63 & 58.23 \\
     & One & CenterPoint-PointPillar~\citep{yin2021center}          & 31.13 & 26.04 & 42.92 & \textbf{23.90} & 19.14 & 50.03 & 60.70 \\
     & One & PillarNet-18~\citep{shi2022pillarnet}                  & \textbf{27.72} & \textbf{25.20} & \underline{28.93} & \underline{24.67} & \underline{19.11} & 59.90 & 67.39 \\
     & One & PillarNeXt~\citep{li2023pillarnext}                    & 28.60 & \underline{25.60} & \textbf{28.50} & 25.10 & 19.20 & \underline{62.20} & \underline{68.40} \\
   
     \cmidrule(rl){2-10}
     & Two & Ours                                                  & \underline{28.41} & 25.66 & 29.32 & 24.96 & \textbf{18.78} & \textbf{63.43} & \textbf{68.96} \\

	 \hline
	 \end{tabular}
	 \end{footnotesize}
	 \end{table*}

	 \vspace{0.3cm}
   \subsection{Discussion}\label{sec:discussion}  

   \setlength{\tabcolsep}{0.2em}
   \begin{table}[t]
   % \captionsetup{font={small}}
   \centering
   % \begin{center}
   \caption{Quantitative comparison for variants of our model. We compute mAP(\%) with 40 recall positions and measure runtime~(Hz) on the validation split of KITTI~\citep{geiger2012we}. 2nd: Two-stage pipeline, S$^{2}$F: Sparse scene feature, CRoIF: Context-aware RoI feature.}
	 %  \begin{adjustbox}{width=1\columnwidth,center}
	 \begin{footnotesize}
		 \begin{tabular}{c c c c | c c c c}
		 \hline
		 SVFM & 2nd & S$^{2}$F & CRoIF & Car & Cyclist & Pedestrian & Speed~(Hz) \\
		 \hline\hline
				 &        &        &        & 77.97  & 62.67 & 53.24 & 39.5 \\
		  \cmark &        &        &        & 82.68  & 66.87 & 60.84 & 35.9 \\
		  \cmark & \cmark &        &        & 83.40  & 69.73  &65.08 & 31.9 \\
		  \cmark & \cmark & \cmark &        & 85.03  & 72.31 & 67.81 & 31.5 \\
		  \cmark & \cmark & \cmark & \cmark &85.57  & 74.87 & 69.45 & 29.6 \\
		 \hline
		 \end{tabular}
		\end{footnotesize}
	 %  \end{adjustbox}
	  \label{table:ablation}
   % \end{center}	
	  \vspace{-0.3cm}
   \end{table}

   \vspace{0.1cm}
	 \subsubsection{Ablation study}
	 We show in Table~\ref{table:ablation} an ablation analysis of each component in our model. We report the mAP performance of three classes,~\ie, car, cyclist, and pedestrian, with 40 recall positions on the moderate split of the validation split of KITTI~\citep{geiger2012we}. We reproduce PointPillars~\citep{lang2019pointpillars} and use it as our baseline in the first row. 
	 % \todo{clarify the experimental configuration for the second model} 
	 From the first and second rows, we can see that SVFM exploiting a stack of pseudo images boosts the detection performance significantly, even without using 3D convolutions, offering mAP gains of $4.04$, $2.91$ and $9.06$ for the car, cyclist and pedestrian classes, respectively. This indicates that preserving 3D structures and leveraging view-specific features are important for accurate 3D object detection, and SVFM extracts discriminative 3D features using 2D convolutions. Our model in the third row adopts a two-stage pipeline without S$^{2}$CFM, that is, we pool RoI features from the stack of pseudo images only. The model in the fourth row, on the other hand, exploits a sparse scene feature for RoI pooling, allowing to boost the detection performance drastically. This suggests that richer contextual information captured by the backbone network is crucial for precise 3D object detection. The average runtime of theses models is almost the same, indicating that S$^{2}$CFM aggregates multi-scale features efficiently. In the last row, the context-aware RoI feature further improves the detection performance for all classes.  We can also observe that the improvements from the context-aware RoI feature are more significant for the cyclist and pedestrian classes. This confirms that leveraging a global context of a 3D scene is particularly effective for small objects captured with sparse points. 
	 % \todo{the final model is more efficient than the third or fourth models? why?}
	 %the small objects with sparse points The improvements are more significant for the cyclist and pedestrian classes than the one for the car class.
	 % \todo{Our model in the third row adopts a two-stage pipeline without S$^{2}$CFM. That is, we pool RoI features from the stack of pseudo images only, while the model in the fourth row exploits a sparse scene feature which boosts the detection performance drastically to extract the RoI features.}

	 \setlength{\tabcolsep}{0.2em} % table 7
	 \begin{table}[!t]
	 % \captionsetup{font={small}}
	 \centering
	 % \begin{center}
	 \caption{Quantitative comparison for SVFMs. We compute mAP(\%) with 40 recall positions on the KITTI~\citep{geiger2012we} validation set. SEQ: Sequential, PAR: Parallel, SEQ-PAR: Sequential-parallel, PAR-SEQ: Parallel-sequential }
	   %  \begin{adjustbox}{width=0.8\columnwidth,center}
	   \begin{footnotesize}
		   \begin{tabular}{C{3em} C{3em} C{3em} C{3em} | c c c }
		   \hline
		   SEQ & PAR & SEQ-PAR & PAR-SEQ & Car & Cyclist & Pedestrian \\
		   \hline\hline
		   \cmark&  &        &  & 85.57  & 74.87 & 69.45 \\
			   &  \cmark       &          &     & 85.32  & 72.46 & 66.98  \\
			&       &    \cmark        &   & 85.49  & 74.65 & 68.65  \\
			&  &            &   \cmark  & 85.28  & 73.82 & 65.68  \\
		   \hline
		   \end{tabular}
		\end{footnotesize}
	   %  \end{adjustbox}
		\label{table:svfm}
	 % \end{center}	
		
	 \end{table}
  
	 \setlength{\tabcolsep}{0.2em} % table 8
	 \begin{table}[!t]
	 % \captionsetup{font={small}}
	 \centering
	 % \begin{center}
	 \caption{Quantitative comparison for various combinations of sparse scene feature~(S$^{2}$F). We compute mAP(\%) with 40 recall positions on the \textit{KITTI}~\citep{geiger2012we} validation set. PI: Pseudo images}
	   %  \begin{adjustbox}{width=0.8\columnwidth,center}
	   \begin{footnotesize}
		   \begin{tabular}{C{3em} C{3em} C{3em} C{3em} | c c c }
		   \hline
		   PI & 1st & 2nd & 3rd & Car & Cyclist & Pedestrian \\
		   \hline\hline
		   \cmark&  &        &  & 83.40  & 69.73  &65.08 \\
		   \cmark   &  \cmark       &          &     & 85.13  & 72.59 & 67.84  \\
		   \cmark&       &    \cmark        &   & 85.46  & 72.38 & 68.54  \\
		   \cmark&  &            &   \cmark  & 83.68  & 73.49 & 68.03  \\
		   \cmark&  \cmark &     \cmark   &  & 85.19  & 72.25 & 67.61 \\
		   \cmark   &  \cmark       &          &   \cmark  & 85.25  & 73.16 & 68.87  \\
		   \cmark&       &    \cmark        & \cmark  & 85.06  & 73.79 & 68.77  \\
		   & \cmark &     \cmark       &   \cmark  & 85.45  & 74.08 & 69.12  \\
		   \cmark& \cmark &     \cmark       &   \cmark  & 85.57  & 74.87 & 69.45  \\
		   \hline
		   \end{tabular}
		\end{footnotesize}
	   %  \end{adjustbox}
		\label{table:sparse}
	 % \end{center}	
	 \end{table}	 

	 \setlength{\tabcolsep}{0.2em} % table 9
	 \begin{table}[!t]
	 % \captionsetup{font={small}}
	 \centering
	 % \begin{center}
	 \caption{Quantitative comparison for different global context features. We compute mAP(\%) with 40 recall positions on the validation split of KITTI~\citep{geiger2012we}. GAP: Global average pooling, S$^{2}$F: Sparse scene feature, Memory: Memory module.}
	   %  \begin{adjustbox}{width=0.8\columnwidth,center}
	   \begin{footnotesize}
		   \begin{tabular}{C{3em} C{3em} C{5em} | c c c }
		   \hline
		   GAP & S$^{2}$F & Memory & Car & Cyclist & Pedestrian \\
		   \hline\hline
		   &  &        & 85.03  & 72.31 & 67.81 \\
		   \cmark     &        &            & 85.12 & 71.21 & 67.45  \\
			   &  \cmark     &              & 85.17 & 73.50 & 68.38 \\
			& \cmark &    \cmark         & 85.57  & 74.87 & 69.45  \\
		   \hline
		   \end{tabular}
		\end{footnotesize}
	   %  \end{adjustbox}
		\label{table:context}
	 % \end{center}	
		\vspace{-0.4cm}
	 \end{table}

	 \vspace{0.1cm}
	 \subsubsection{SVFM}
	 In Table~\ref{table:svfm}, we compare the quantitative results with various backbone networks using different versions of SVFM~(Fig.\ref{fig:svfm_blocks}). We train all models with the KITTI train set, and report their performances in terms of mAP for three classes,~\ie, car, cyclist, and pedestrian, on the moderate split of the KITTI validation set. We observe that SVFMs with sequential structures in the first and the third rows provide better detection results for all classes than the parallel ones, and the parallel SVFM in the second row gives the worst result. This suggests that capturing correlations among view-specific features is crucial for learning discriminative 3D voxel features. 
	 % \todo{clarify: sequential structures can capture correlations? why?}
   %  Based on the results, we use the sequential SVFM for all experiments, and also report the results of our model with the sequential-parallel SVFM for comparisons with other methods.  

   %  , where the RoI features are pooled from. We train all models with the \textit{KITTI} train set. We compare the performance of each model in terms of mAP, and report results of three classes,~\ie, car, cyclist, and pedestrian, on the moderate split of the \textit{KITTI} validation set. Our full model's results are presented in the last row for comparison. 

   \vspace{0.1cm}
   \subsubsection{Sparse scene feature}

	 We show in Table~\ref{table:sparse} quantitative results for various combinations of sparse scene feature. From the table, we can observe three things: (1) Our models using multi-scale features consistently outperform the one without the features~(the 1st row). This indicates that leveraging rich contexts learned from the multi-scale features are effective for accurate 3D object detection. (2) Exploiting all levels of features boosts the detection performance drastically~(the 8th and last rows), showing that they are complementary each other. (3) Each object class benefits differently from multi-scale features~(from the 2nd to 4th rows), with the first-level feature offering the best improvement for the car class, while the second- and third-level features enhance the detection performance for the pedestrian and cyclist classes, respectively. This emphasizes the importance of integrating all features of various scale for precise localization of diverse objects. 
   
   %We compare the performance of each model in terms of mAP, and report results of three classes,~\ie, car, cyclist, and pedestrian, on the moderate split of the KITTI validation set. For comparison with our final model in the last row, we compute other context-aware RoI features in two ways:

	 \subsubsection{Global context feature}
	 We show in Table~\ref{table:context} quantitative results for different types of global context features used to obtain a context-aware RoI feature. To this end, we augment the sub-RoI features in two other ways: (1) We represent global contextual information as a single feature vector by applying a global average pooling~(GAP) to a sparse scene feature, and concatenate the result with each sub-RoI one~(the 2nd row). (2) We directly use the sparse scene feature as the global context feature without a memory module, which integrates the sparse scene and sub-RoI features~(the 3rd row). Specifically, we compute matching probabilities between the sparse scene and sub-RoI features, and aggregate the scene features with corresponding probabilities for each sub-RoI feature, which is analogous to the aggregated context features in Eq.~\eqref{eq:wg}. In the first row, we show the results of our model without a context-aware RoI feature. The result in the second row shows that a single context feature could not improve the detection performance across all classes. On the contrary, we can observe that the performance gains from incorporating multiple global context features,~\ie, exploiting the sparse scene feature, are more significant than those from a single context feature. This indicates that representing global contexts as a single feature vector may not be enough to capture the contextual information in complex outdoor scenes. We can also see that the performance improvements are more significant for the cyclist and pedestrian classes, suggesting that augmenting the RoI features with global context features is particularly effective for small objects with sparse points. In the last row, the memory module further boosts the detection performance for all classes, confirming that richer contextual information from diverse but similar scenes is crucial for outdoor scenes.

	 \begin{figure}[!t] %figure 6
		% \captionsetup{font={footnotesize}}
	   \centering
	   \includegraphics[width=1.0\linewidth]{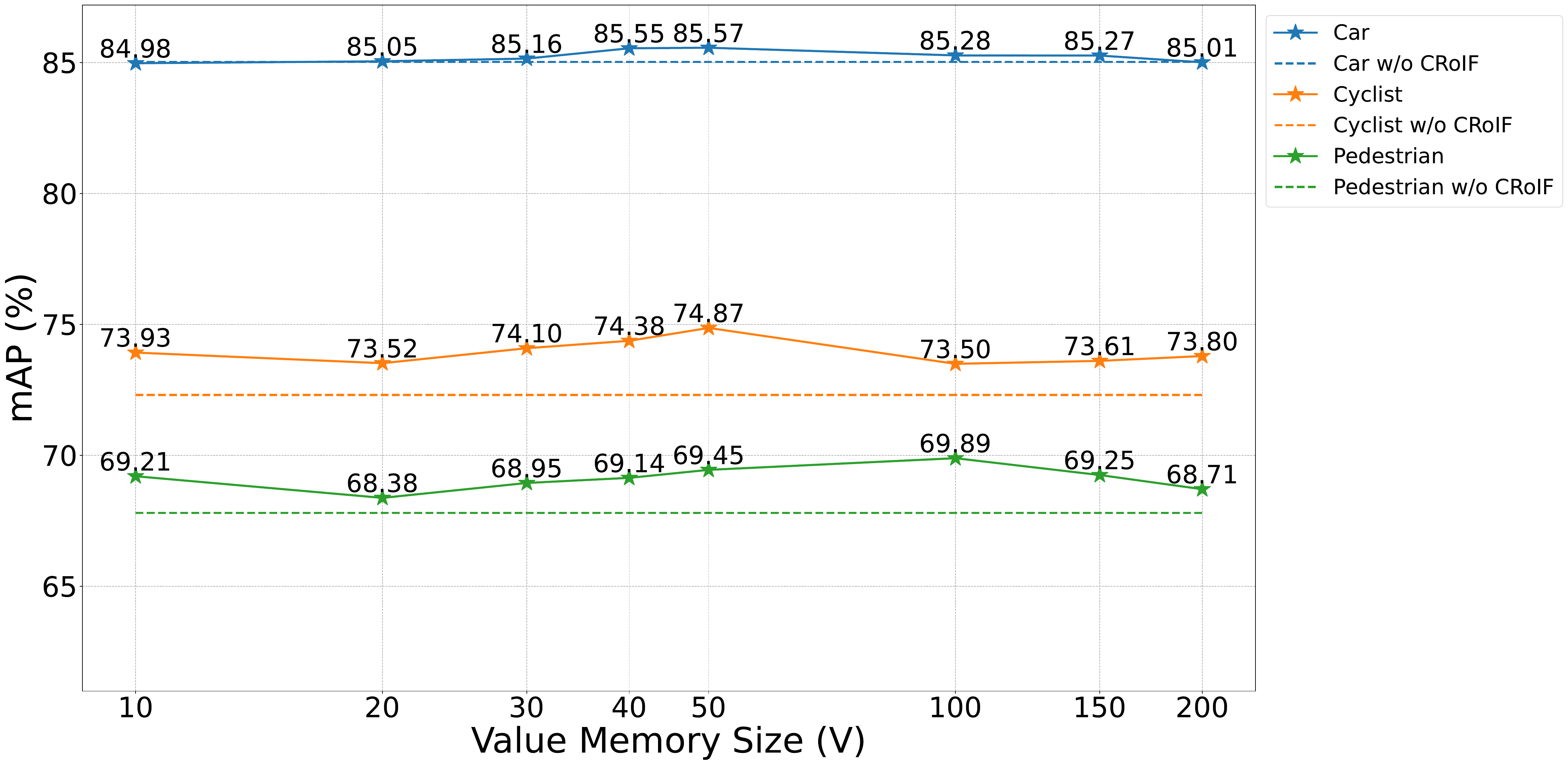}
		  \caption{Quantitative comparison according to the size of a value memory~($V$). We compute mAP(\%) with 40 recall positions on the validation split of KITTI~\citep{geiger2012we}.}
		  \label{fig:memory}
		  \vspace{-0.4cm}
	   \end{figure}
	 
	   \setlength{\tabcolsep}{0.2em}
	   \begin{table}[!t]
		 % \captionsetup{font={small}}
		 \centering
		 \caption{Comparison of the number of network parameters.}
		 % \begin{adjustbox}{width=0.8\columnwidth,center} 
		   \centering
		   \begin{footnotesize}
		   \begin{tabular}{c c|c}
		   \hline
			Stage & Methods        & Number of parameters \\ \hline
			One&PointPillars~\citep{lang2019pointpillars}      & \hspace{0.1cm}$4.8$M  \\
			One&SECOND~\citep{yan2018second}      & \hspace{0.1cm}$5.3$M  \\
			Two&PV-RCNN~\citep{shi2020pv} & $13.1$M  \\ 
			Two&Voxel R-CNN~\citep{deng2021voxel}  & \hspace{0.1cm}$9.8$M \\
			Two&CT3D~\citep{sheng2021improving} & $30.0$M \\
			Two&VoTR-TSD~\citep{mao2021voxel}& $12.6$M \\
			Two&Focals Conv~\citep{chen2022focal}& $13.4$M \\
			One&PillarNet-18~\citep{shi2022pillarnet}  & $11.0$M   \\
			Two&PV-RCNN++~\citep{shi2023pv} & $14.3$M  \\ 
			One&PillarNext-B~\citep{li2023pillarnext} & \hspace{0.1cm}$6.9$M \\
			 \hline
			Two &Ours    &  \hspace{0.1cm}$8.1$M \\ \hline
		   \end{tabular}
		 \end{footnotesize}
		 % \end{adjustbox}
		 \label{tab:param}
		 \vspace{-0.2cm}
	   \end{table}

	   \setlength{\tabcolsep}{0.2em}
	 \begin{table}[!t]
		   % \captionsetup{font={small}}
		   \centering
		   \caption{Runtime comparison of our model and PointPillars~\citep{lang2019pointpillars}.}
		   % \begin{adjustbox}{width=0.8\columnwidth,center} 
			 \centering
			 \begin{footnotesize}
			 \begin{tabular}{c|c c}
			 \hline
			 \multirow{2}{*}{Step}        & Ours & PointPillars~\citep{lang2019pointpillars} \\ 
			 & Time~(ms) & Time~(ms) \\ \hline
			 Pseudo images & \hspace{0.1cm}$4.2$ & $3.9$ \\ 
			 Backbone  & \hspace{0.1cm}$8.9$  & $9.3$\\
			 RPN      & \hspace{0.1cm}$4.6$ &$4.4$  \\
			 RoI head           & $12.4$ & -  \\ 
			 Post-processing  & \hspace{0.1cm}$3.7$ & $6.3$ \\ \hline
			 \end{tabular}
		   % \end{adjustbox}
			\end{footnotesize}
		   \label{tab:runtime}
		   \vspace{-0.2cm}
		\end{table}

	 \subsubsection{Memory size}
	 We show in Fig.~\ref{fig:memory} quantitative results of our model with varying the size of a value memory~$V$. We evaluate our model on the moderate split of the KITTI validation set, and compare the performance in terms of mAP. The size of key memory~($K$) is fixed to~$10$. We can observe that using more value items for storing various concepts of the prototypical contexts generally provides better results. However, exploiting too many items rather degrades the results, suggesting that setting an appropriate memory size is crucial for the detection performance. We can also see that that all models with a memory module consistently outperform the ones without the memory.

	 % \subsubsection{Key-memory distribution.}
	   % 	\todo{memory ablation orthogonal loss, We visualize in Fig.~\ref{fig:tSNE} the distribution of query features of our model for the reconstruction task, randomly chosen from UCSD Ped2~\citep{li2013anomaly}, learned with and without the  feature separateness loss. We can see that our model trained without the separateness loss lose the discriminability between items in the memory, and thus all features are mapped closely in the embedding space. The separateness loss allows to separate individual items in the memory, demonstrating that it enhances the discriminative power of query features and memory items significantly. We can also see that our model gives compact feature representations.}  
   
	   \begin{figure*}[!t] 
		% \captionsetup[subfigure]{aboveskip=1pt,belowskip=1pt,justification=centering}
		% \captionsetup{font={footnotesize}}
		\begin{adjustbox}{width=2.0\columnwidth,center} 
		  \subfigure[]
			{\includegraphics[width=0.25\textwidth, height=0.5\textwidth]{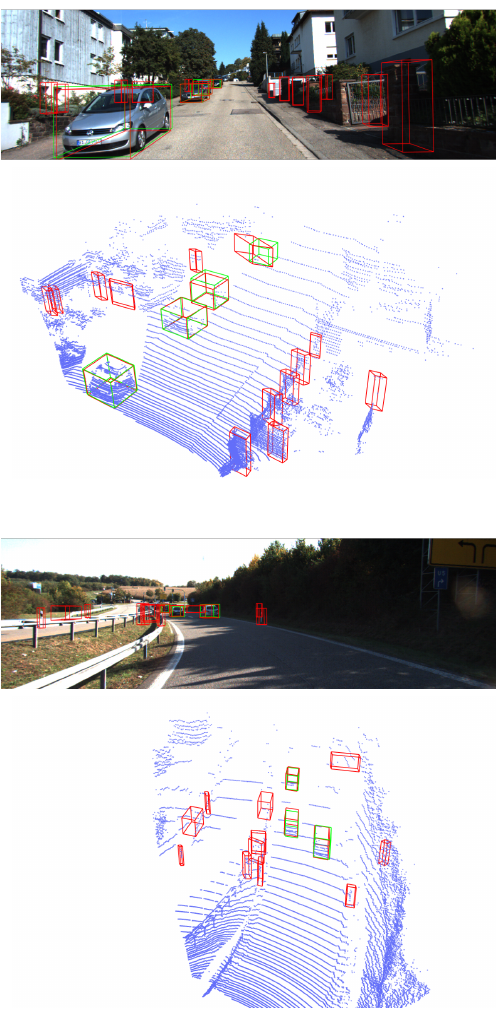}}
		  \subfigure[]
			{\includegraphics[width=0.24\textwidth, height=0.5\textwidth]{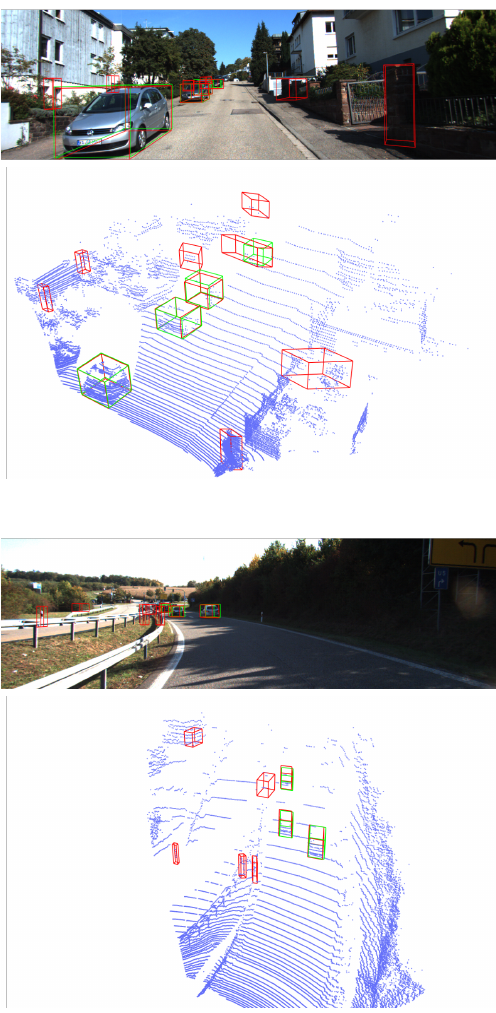}}
		  \subfigure[]
			{\includegraphics[width=0.25\textwidth, height=0.5\textwidth]{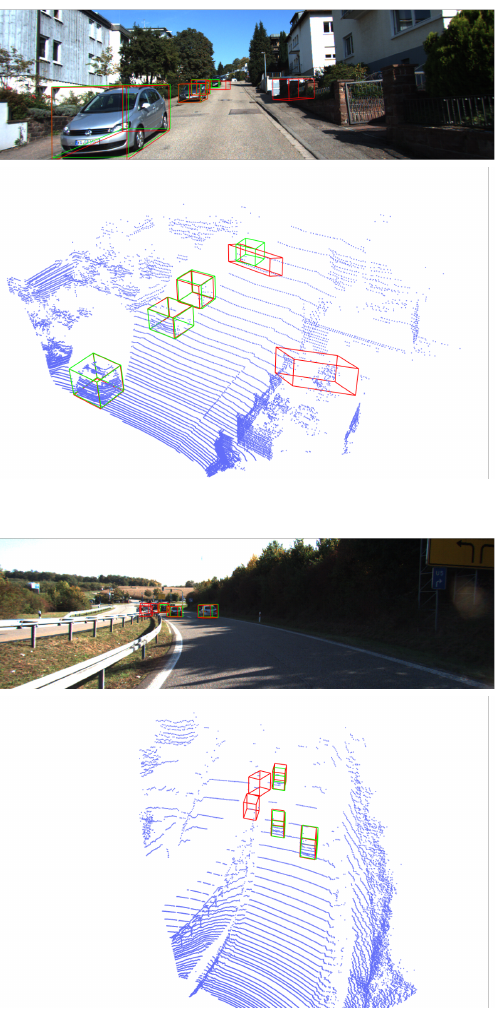}}
		  \subfigure[]
			{\includegraphics[width=0.25\textwidth, height=0.5\textwidth]{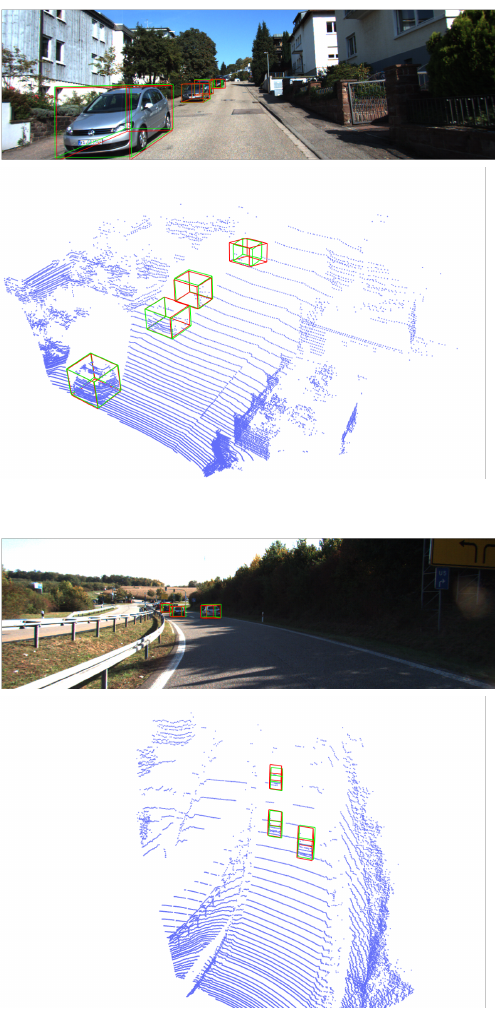}}
		\end{adjustbox}
		\vfill
		\caption{Qualitative results on the validation set of KITTI~\citep{geiger2012we}. We show our predictions and ground-truth bounding boxes as red and green boxes, respectively. In particular, our model with all components localizes small objects well consisting of sparse point clouds. We also visualize 3D bounding boxes in RGB images projected from 3D detection results. CRoIF: Context-aware RoI feature. (a)~Baseline. (b)~Ours w/o RoI head. (c)~Ours w/o CRoIF. (d)~Ours (Best viewed in color.)}
		\label{fig:qualitative}    
		\vspace{-0.7cm}
	  \end{figure*}

	  \begin{figure*}[!t]
		\centering
		
		% --- Row 1 (no subcaption) ---
		\includegraphics[width=0.19\textwidth]{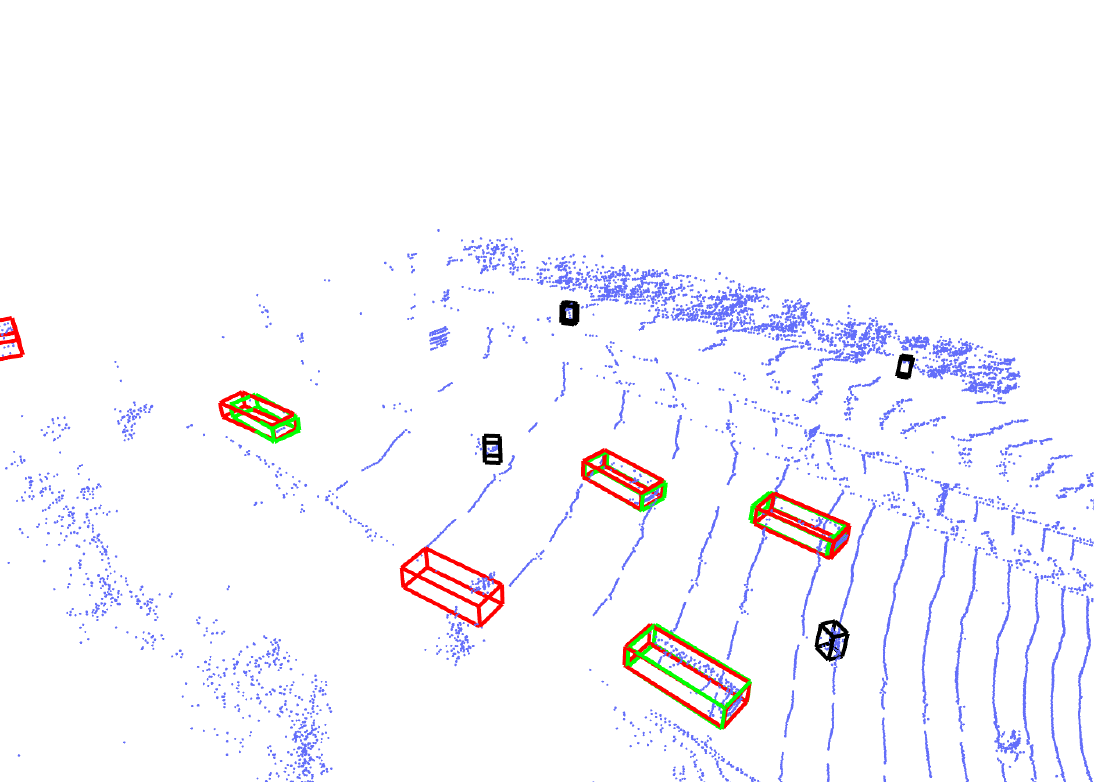}
		\includegraphics[width=0.19\textwidth]{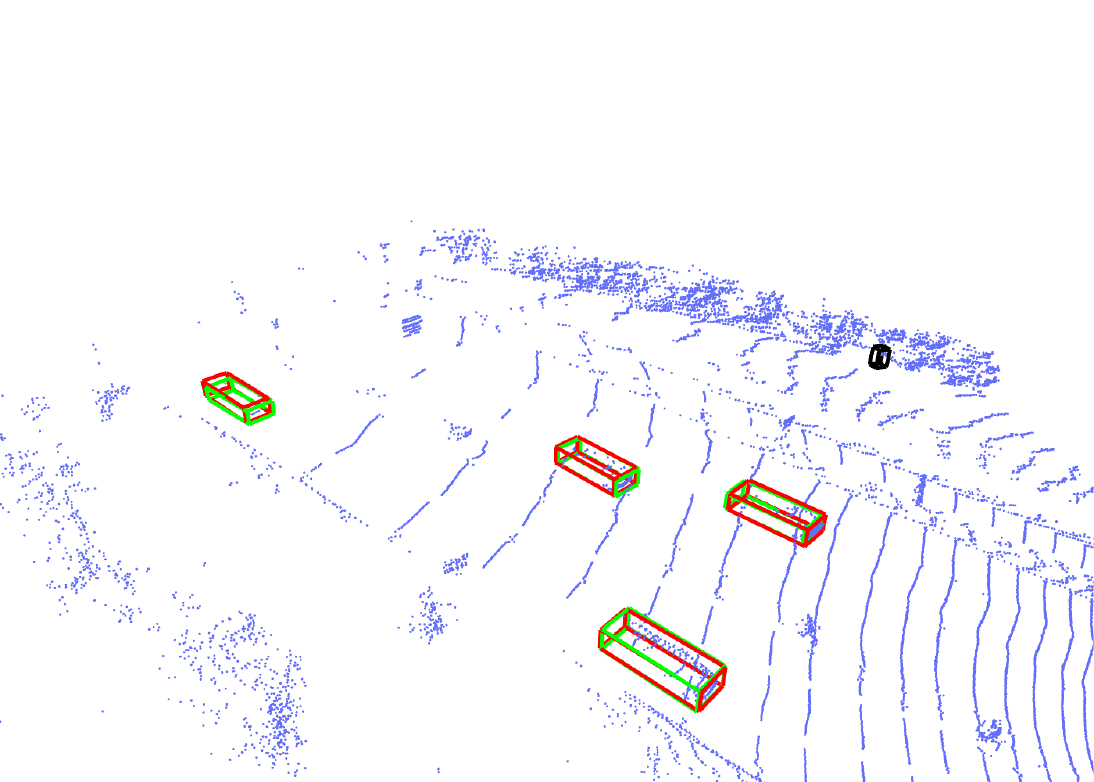}
		\includegraphics[width=0.19\textwidth]{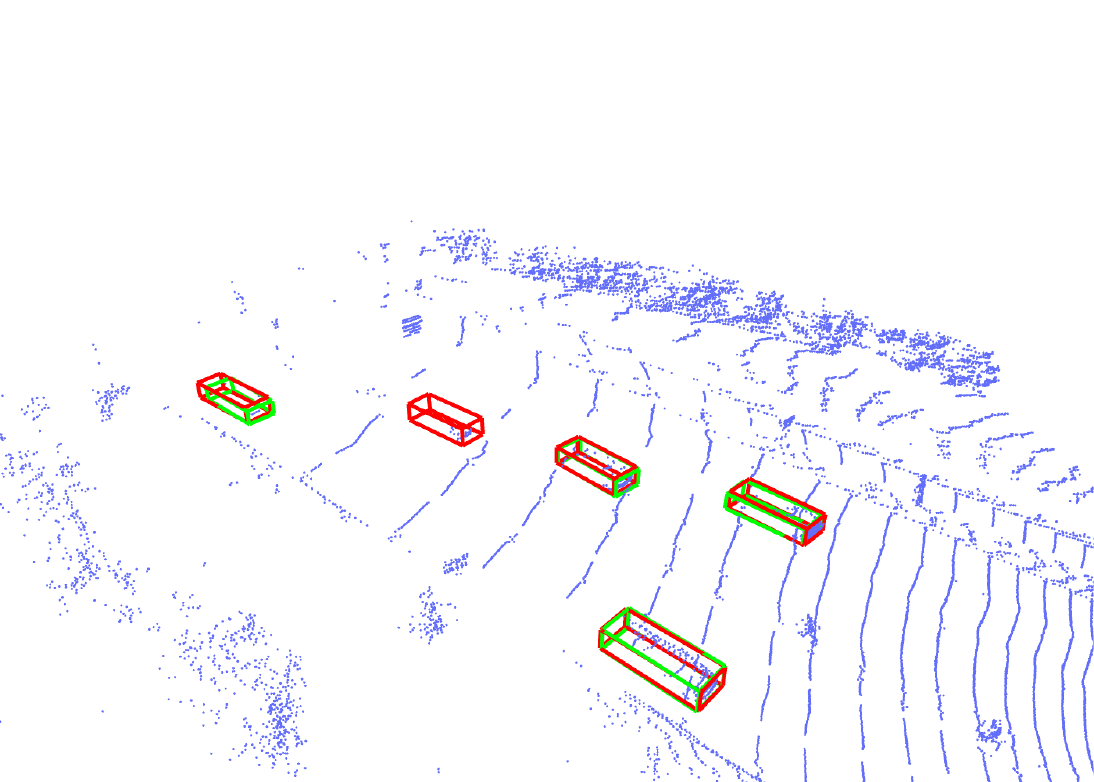}
		\includegraphics[width=0.19\textwidth]{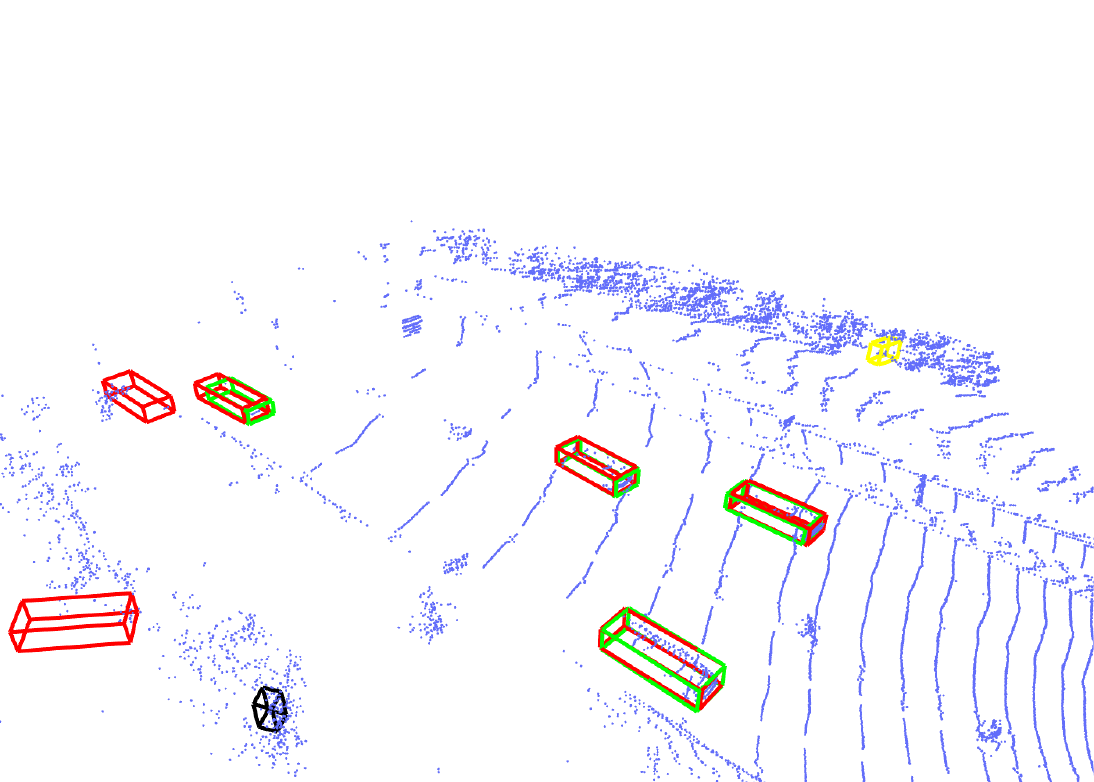}
		\includegraphics[width=0.19\textwidth]{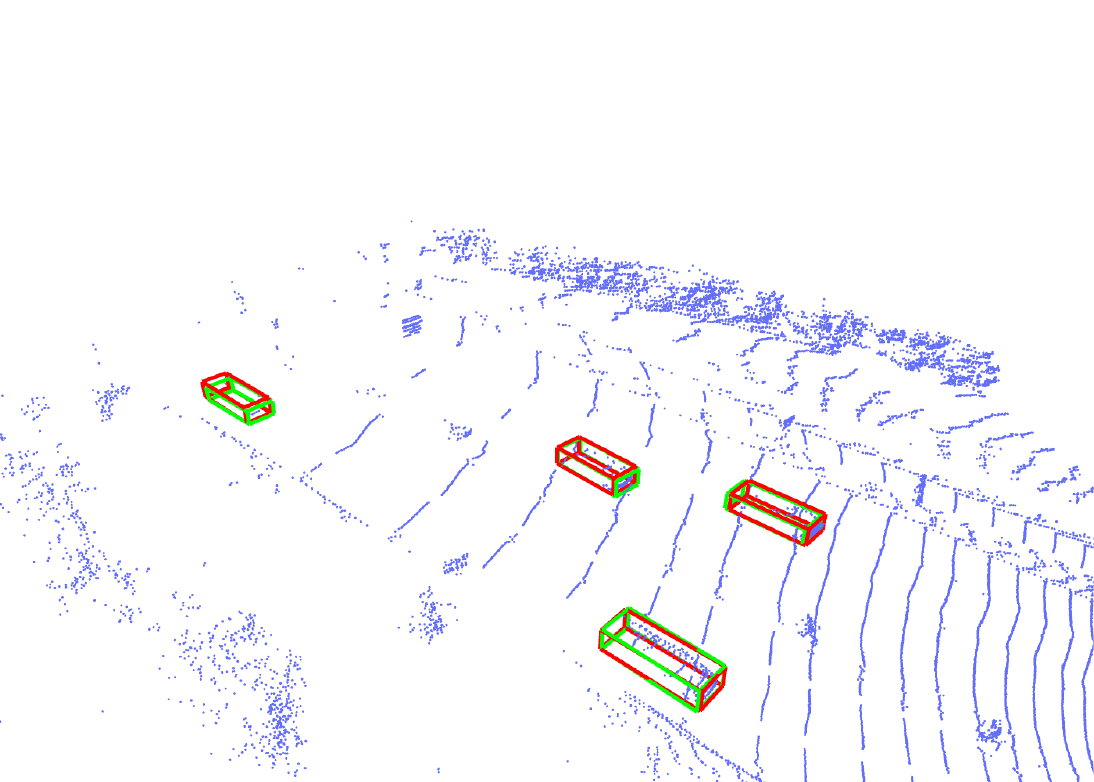} \\
		
		\vspace{1mm}
		
		% --- Row 2 (no subcaption) ---
		\includegraphics[width=0.19\textwidth]{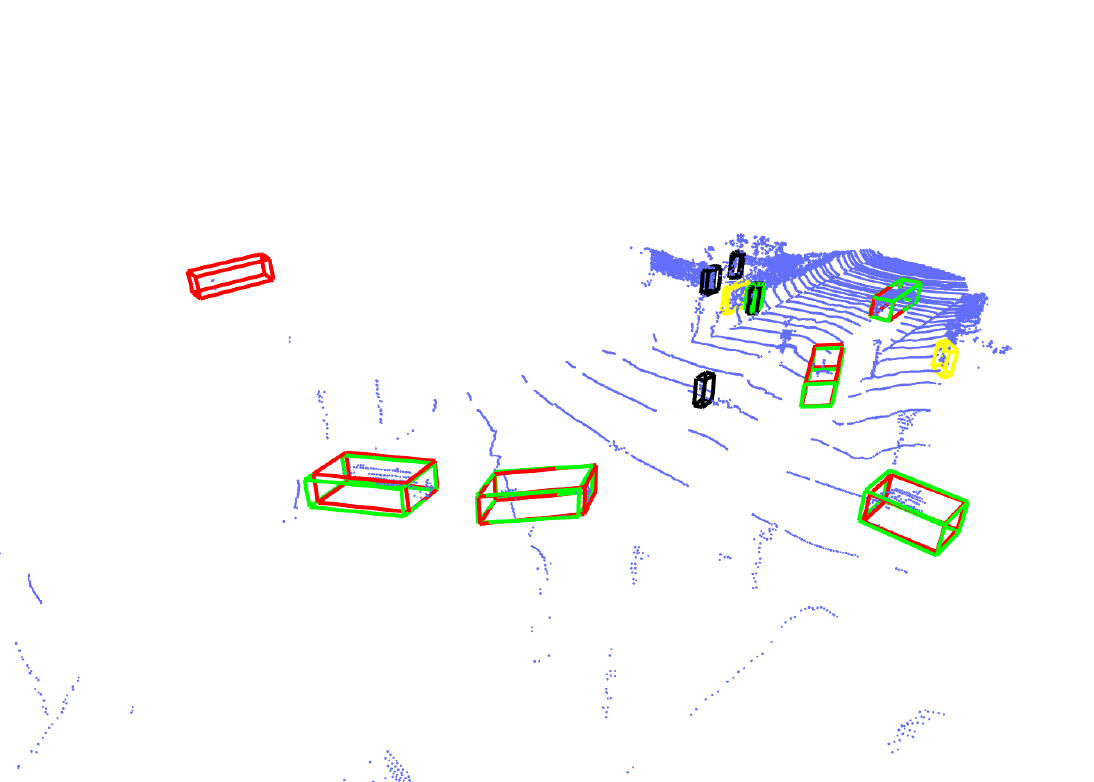}
		\includegraphics[width=0.19\textwidth]{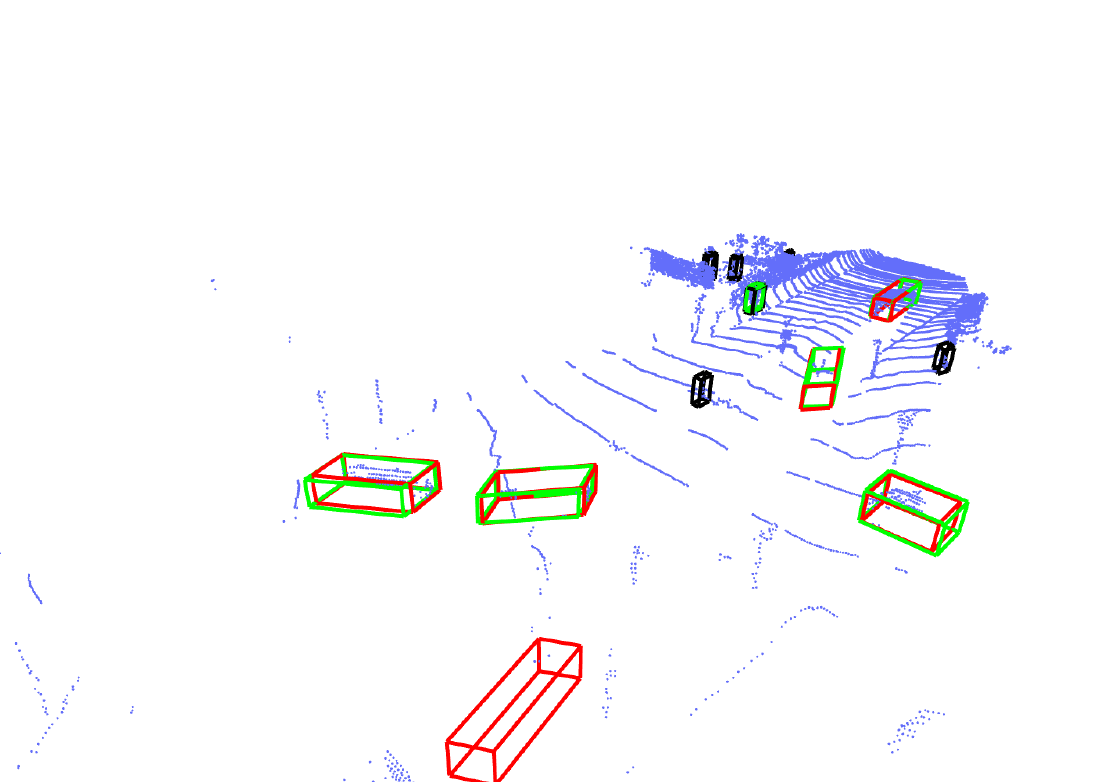}
		\includegraphics[width=0.19\textwidth]{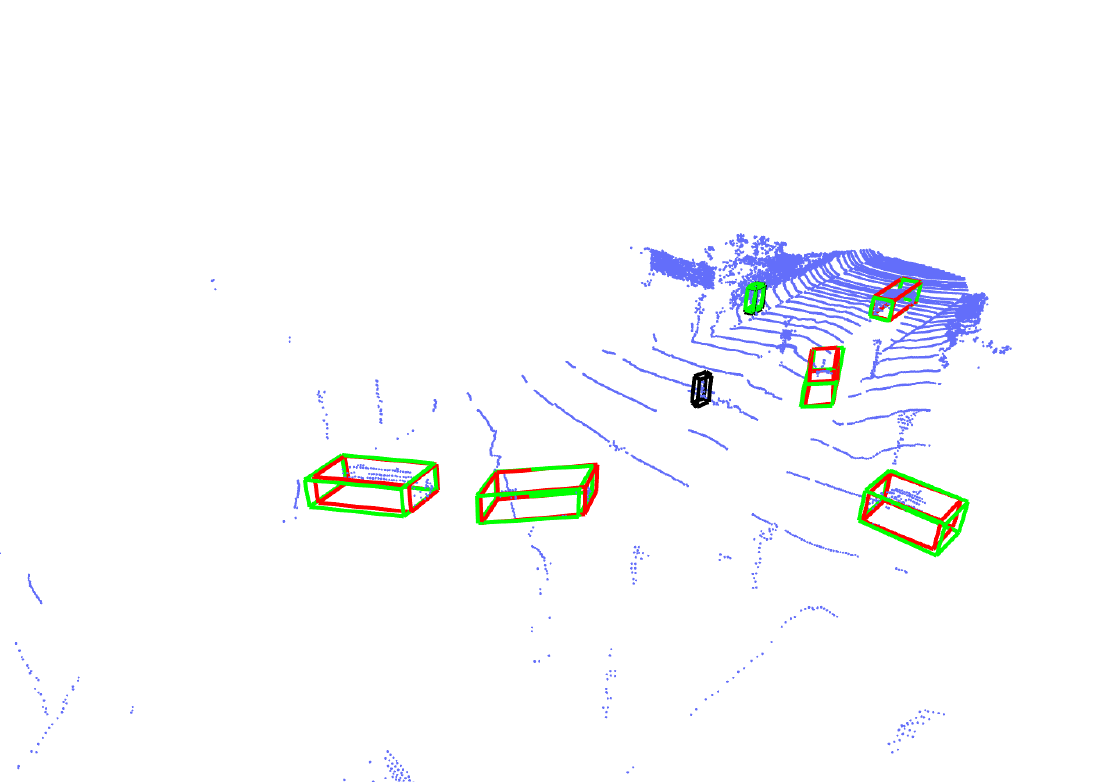}
		\includegraphics[width=0.19\textwidth]{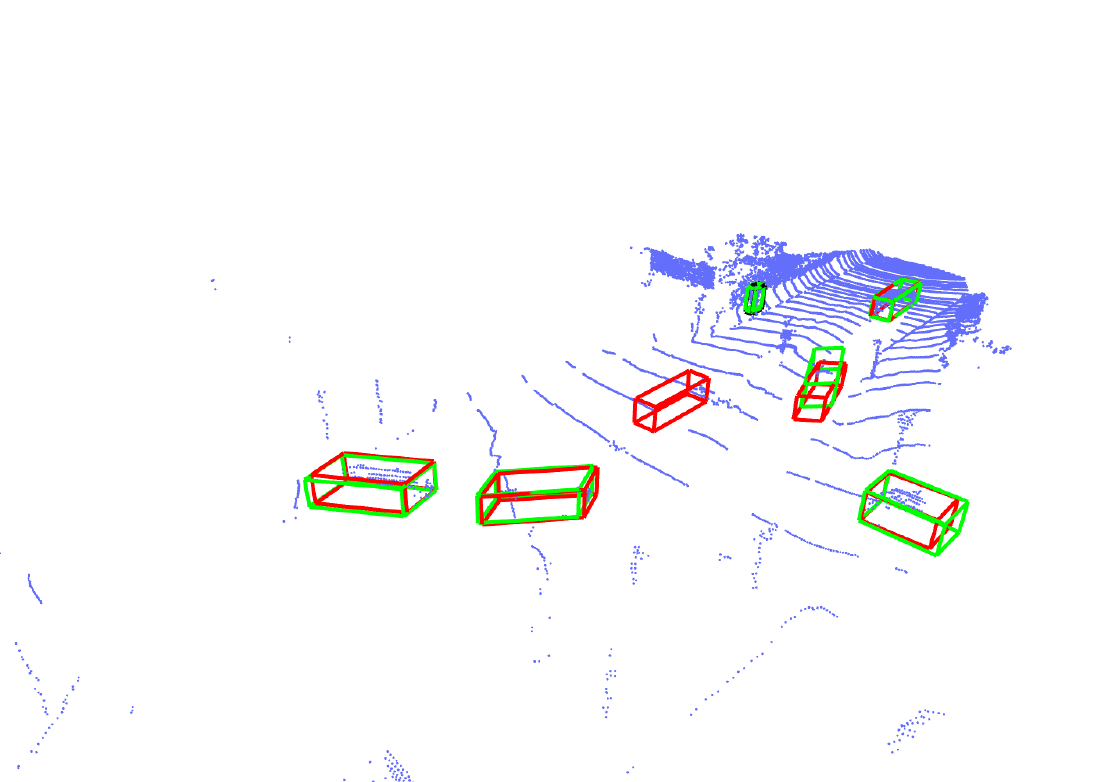}
		\includegraphics[width=0.19\textwidth]{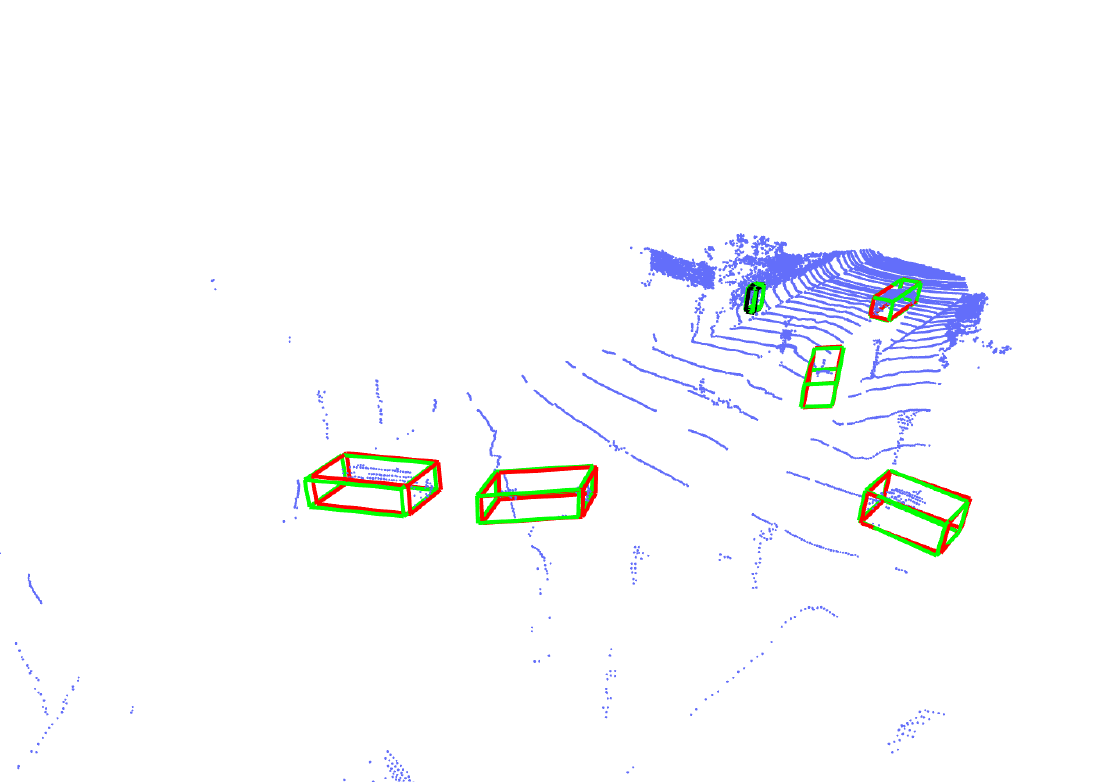}
		
		\vspace{1mm}
		
		% --- Row 3 (with subcaptions) ---
		\subfigure[]{\includegraphics[width=0.19\textwidth]{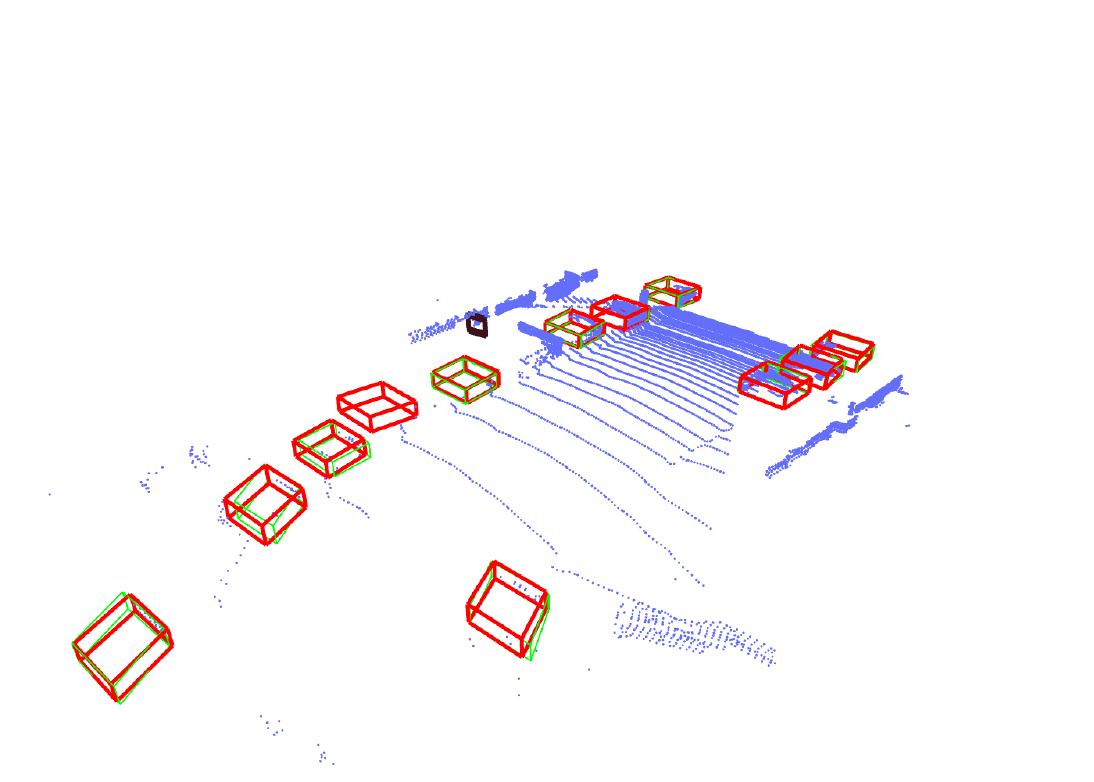}}
		\subfigure[]{\includegraphics[width=0.19\textwidth]{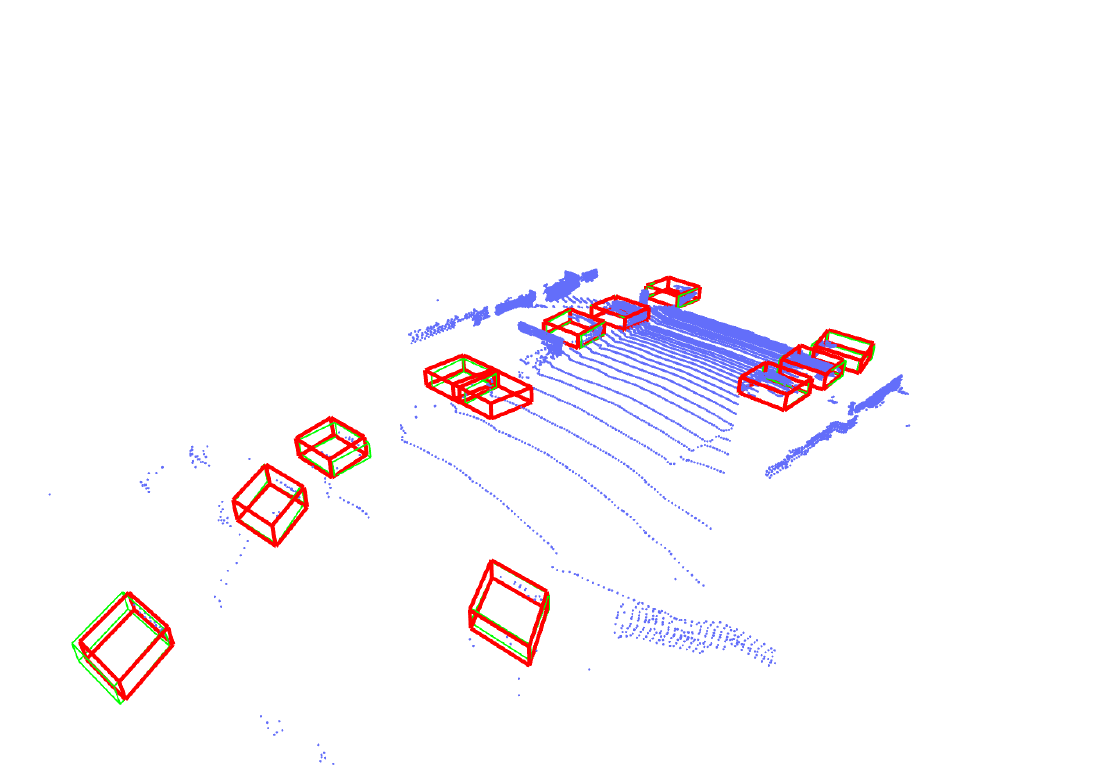}}
		\subfigure[]{\includegraphics[width=0.19\textwidth]{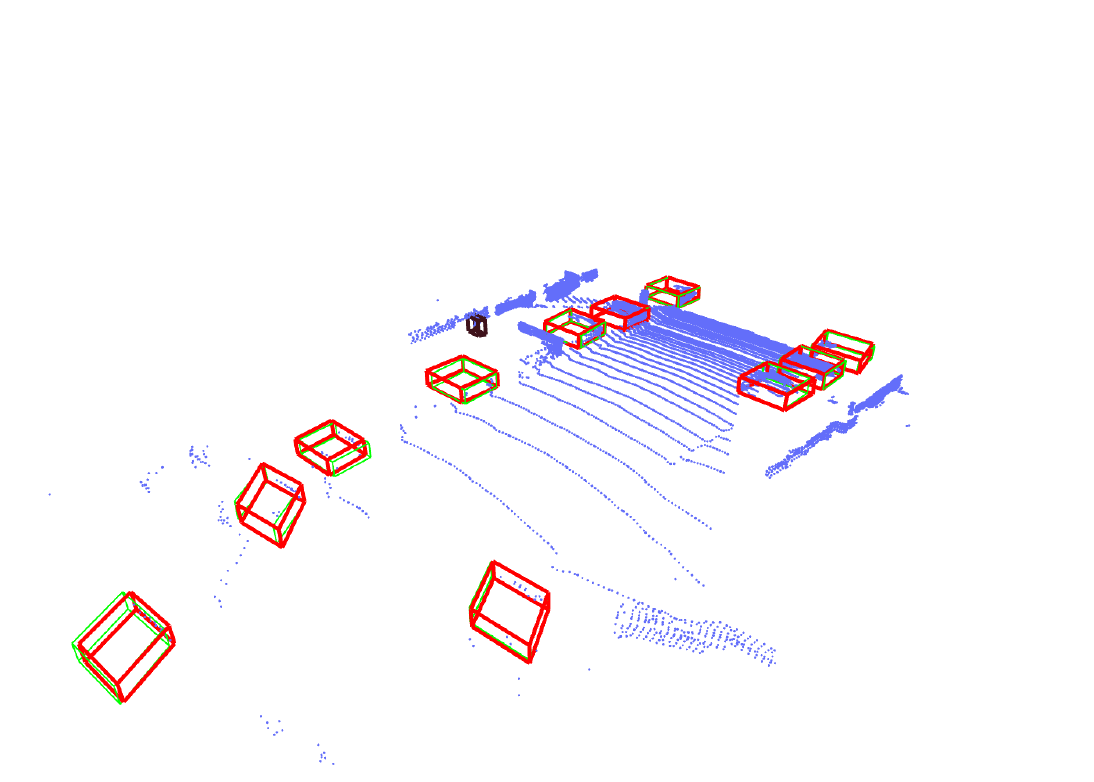}}
		\subfigure[]{\includegraphics[width=0.19\textwidth]{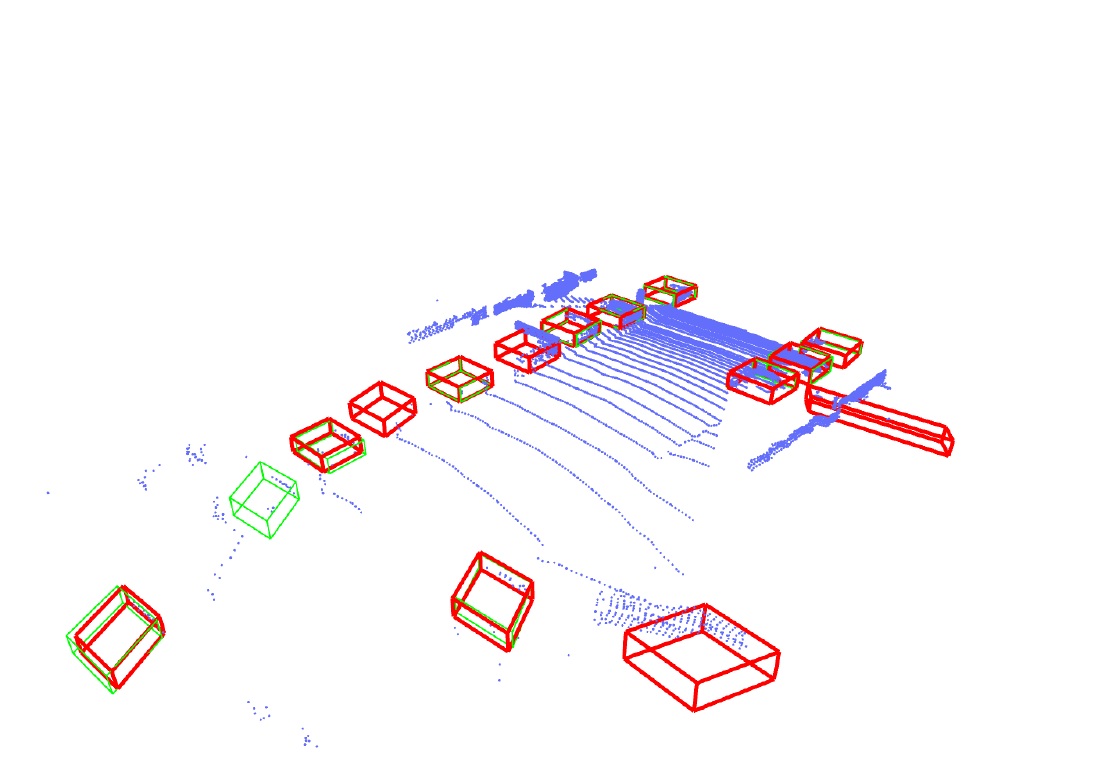}}
		\subfigure[]{\includegraphics[width=0.19\textwidth]{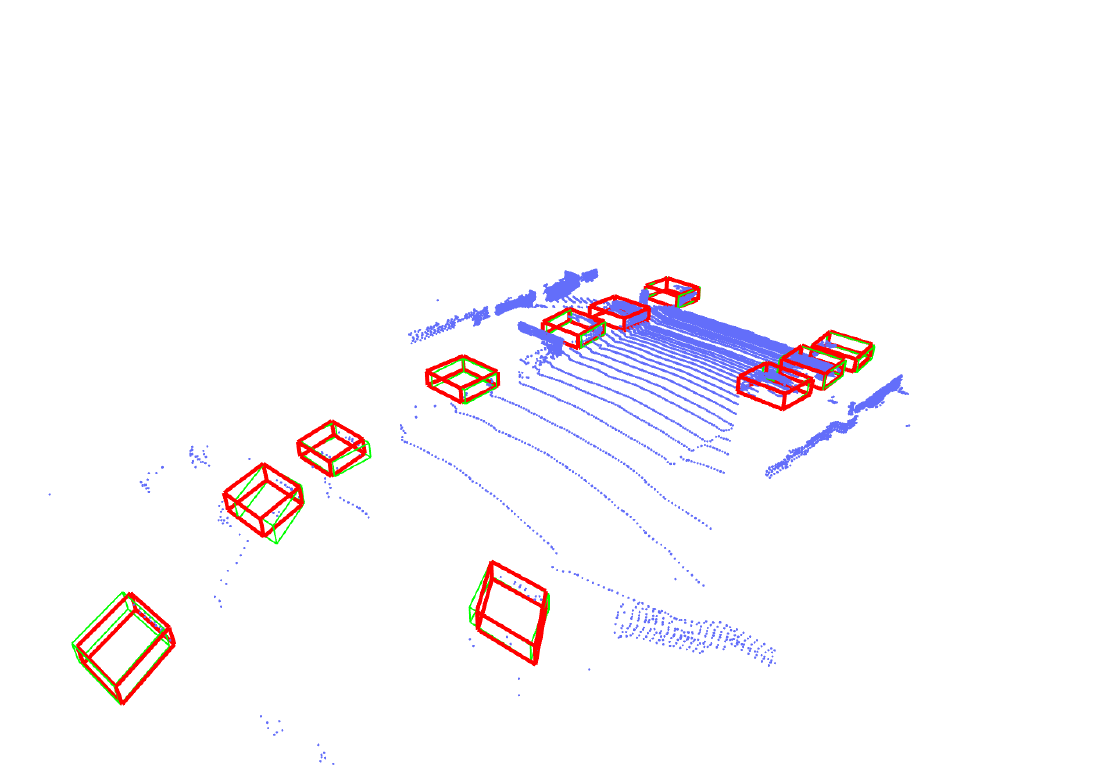}}

		\vspace{-1mm}
		\caption{Qualitative comparison with the state of the art on the validation set of KITTI~\citep{geiger2012we}. We show ground-truth bounding boxes as green boxes, and predictions as red, black, and yellow boxes for car, pedestrian, and cyclist, respectively. We can clearly see that our method provide better results than other methods. (a)~PointPillar~\citep{wang2020pillar}. (b)~PVRCNN~\citep{shi2023pv}. (c)~VoxelRCNN~\citep{deng2021voxel}. (d)~PillarNet~\citep{shi2022pillarnet}. (e)~Ours. (Best viewed in color.)}
		\label{fig:qual_compare}
		\vspace{-0.7cm}
		\end{figure*}
		\begin{figure*}[t]
			\captionsetup[subfigure]{aboveskip=1pt,belowskip=1pt,justification=centering}
			% \captionsetup{font={footnotesize}}
			\begin{adjustbox}{width=2.0\columnwidth,center} 
			  \subfigure
				{\includegraphics[width=0.5\textwidth, height=0.5\textwidth]{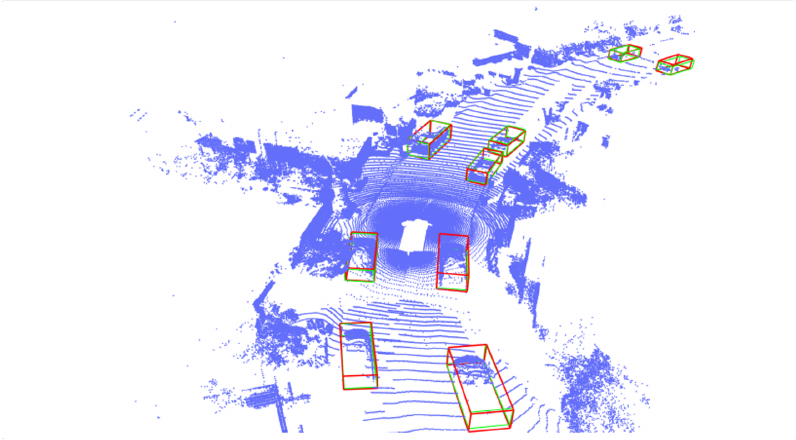}}
			  \subfigure
				{\includegraphics[width=0.5\textwidth, height=0.5\textwidth]{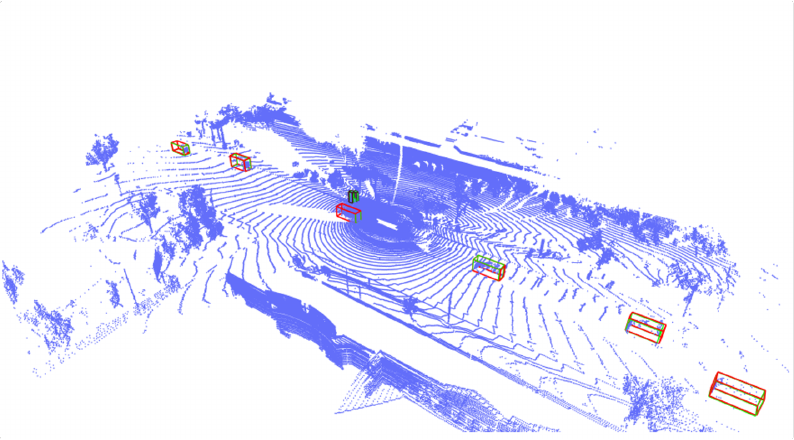}}
			  \subfigure
				{\includegraphics[width=0.5\textwidth, height=0.5\textwidth]{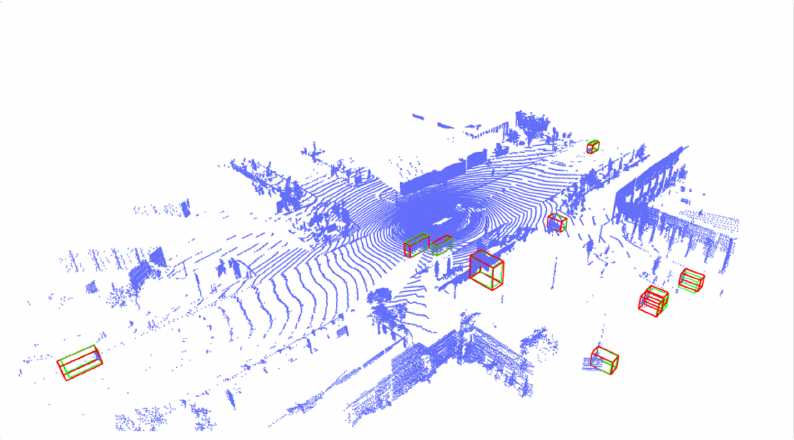}}
			  \subfigure
				{\includegraphics[width=0.5\textwidth, height=0.5\textwidth]{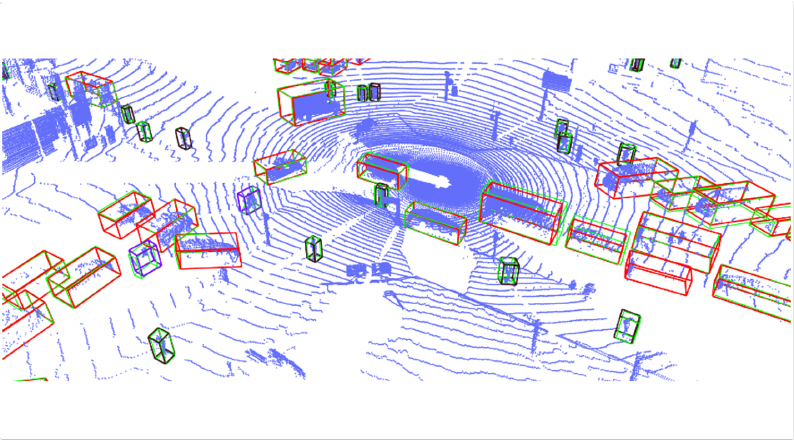}}
			\end{adjustbox}
			\vfill
			\vspace{-0.2cm}
			\caption{Qualitative results on the validation set of the Waymo Open Dataset~\citep{sun2020scalability}. We show ground-truth bounding boxes as green boxes, while our predictions for the vehicle class are represented with red boxes. We visualize predictions of our model for the cyclist and pedestrian classes with magenta and violet boxes, respectively. (Best viewed in color.)}
			\label{fig:quali_waymo}    
			\vspace{-0.6cm}
		  \end{figure*}

	 \subsubsection{Model efficiency}
   We compare in Table~\ref{tab:param} the numbers of network parameters for our model and other detection methods using 3D convolutions~\citep{shi2020pv,deng2021voxel,yan2018second,sheng2021improving,mao2021voxel,chen2022focal,shi2023pv} or 2D convolutions~\citep{lang2019pointpillars,li2023pillarnext,shi2022pillarnet}. We can clearly see that our model uses much fewer parameters than other approaches, except for the single-stage methods of~\citep{yan2018second, lang2019pointpillars, li2023pillarnext}. This indicates that SVFM reduces the number of parameters significantly by replacing the expensive 3D operations with 2D convolutions. In Table~\ref{tab:runtime}, we show an average runtime of a baseline model~(PointPillars~\citep{lang2019pointpillars}) and ours in detail. We split a detection process into five steps and measure the runtime of each step with a Geforce RTX 2080 Ti GPU. We can see that our model spends most computations to refine 3D object proposals in the RoI head, while the runtime of other steps is comparable to that of the baseline. Regarding the added complexity of the second stage, we further analyze the computational and memory overhead introduced by S$^{2}$CFM module in the RoI head. To this end, we compare our full model with a variant that applies the second stage without S$^{2}$CFM (Table~\ref{table:ablation}, third row). We have observed that the additional memory usage does not critically affect the overall efficiency of the model. Specifically, the memory usage increases by only 115MB~(9.22\% of the total usage). The computational overhead is also modest, adding approximately 6ms per forward pass,~\ie, 17.75\% of the total runtime and roughly half of the RoI head's processing time. These overheads primarily stem from (1) memory modules storing key and value items, (2) key addressing in Eq.~\eqref{eq:wk}, which computes cross-attention between key and sub-RoI features, and (3) storing multi-scale features to construct sparse scene features.
   %  The 2-stage approach makes our method slower than PointPillars. 

   % \vspace{-0.05cm}
   
	 \subsubsection{Qualitative results}
	 We show in Fig.~\ref{fig:qualitative} object detection results on the KITTI validation set. We can clearly see that our method shows better results than the baseline. The two-stage models ({Figs.~\ref{fig:qualitative}(c) and~(d)}) provide more reliable predictions than one-stage models~({Figs.~\ref{fig:qualitative}(a) and~(b)}), reducing false positives significantly. Although the two-stage framework in (c) yields satisfactory results, our model with a context-aware RoI feature provides more precise results. For example, it localizes the distant object with sparse points more accurately~(\eg, the farthest car in the first scene), confirming once again the effectiveness of our approach to adopting two-stage strategy.  
	 We also compare in Fig.~\ref{fig:qual_compare} our approach to PointPillar~\citep{lang2019pointpillars}, PVRCNN~\citep{shi2020pv}, VoxelRCNN~\citep{deng2021voxel}, and PillarNet~\citep{shi2022pillarnet} on the validation set of KITTI. We can see that our methods provides better results than other methods. From the figure, we have two findings as follows: (1) Our method detects objects more accurately than other methods, surpressing false positives that often occur in one-stage dtectors~(See Figs.~\ref{fig:qual_compare}(a), (d), and (e)). A plausible reason is that our approach enables adopting two-stage strategy, providing well-aligned and reliable predictions across the scene. (2) SVFM and S$^{2}$CFM work favorably for sparse point distributions, effectively localizing small and distant objects. This indicates that SVFM enhances feature representations by aggregating multi-scale spatial cues, while S$^{2}$CFM refines RoI features using context-aware memory, enabling more precise localization even under sparse point distributions.
	 We also show in Fig.~\ref{fig:quali_waymo} detection results on the validation set of Waymo Open Dataset. We can observe that our method is robust to complex scale variations. For example, it localizes 3D objects well, including distant ones captured with sparse point clouds~(\eg,~vehicle at bottom left corner of the third scene) and small ones like cyclists or pedestrians in the fourth scene. This also confirms the effectiveness of SVFM and S$^{2}$CFM.

	  \begin{figure*}[t]
		\captionsetup[subfigure]{aboveskip=1pt,belowskip=1pt,justification=centering}
		\centering
		% \captionsetup{font={footnotesize}}
		% \begin{adjustbox}{width=1.0\linewidth,center} 
		  \subfigure[]
			{\includegraphics[width=0.48\textwidth, height=0.28\textwidth]{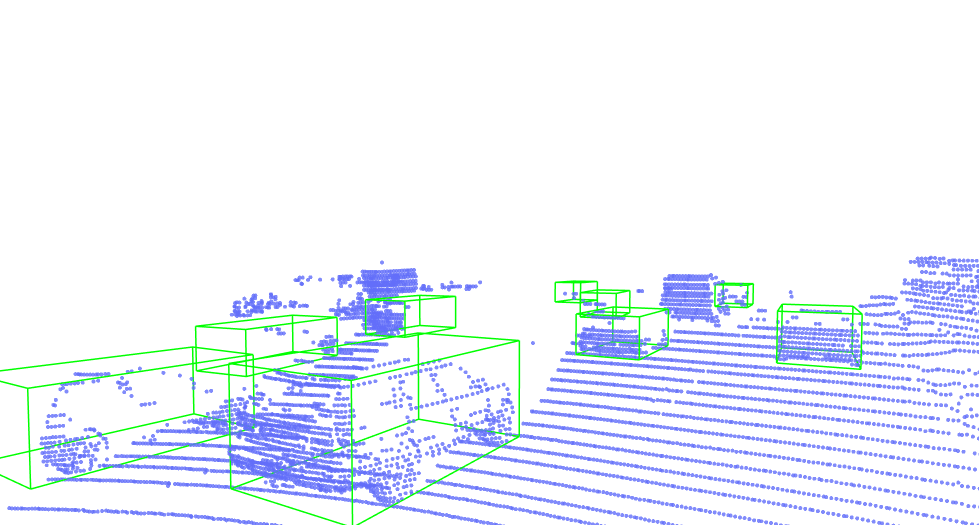}}
		  \subfigure[]
			{\includegraphics[width=0.48\textwidth, height=0.28\textwidth]{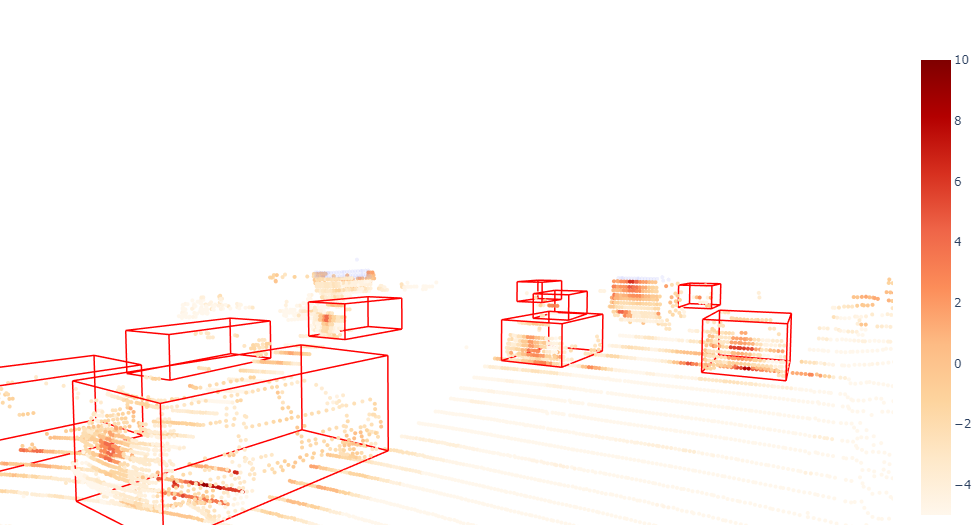}}
			\hspace{-0.3em}
		  \subfigure[]
			{\includegraphics[width=0.48\textwidth, height=0.28\textwidth]{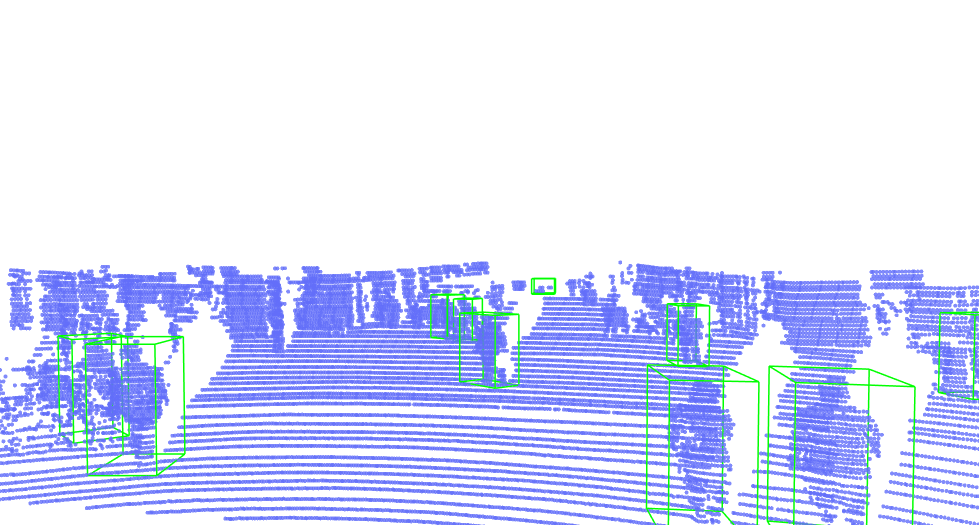}}
		  \subfigure[]
			{\includegraphics[width=0.48\textwidth, height=0.28\textwidth]{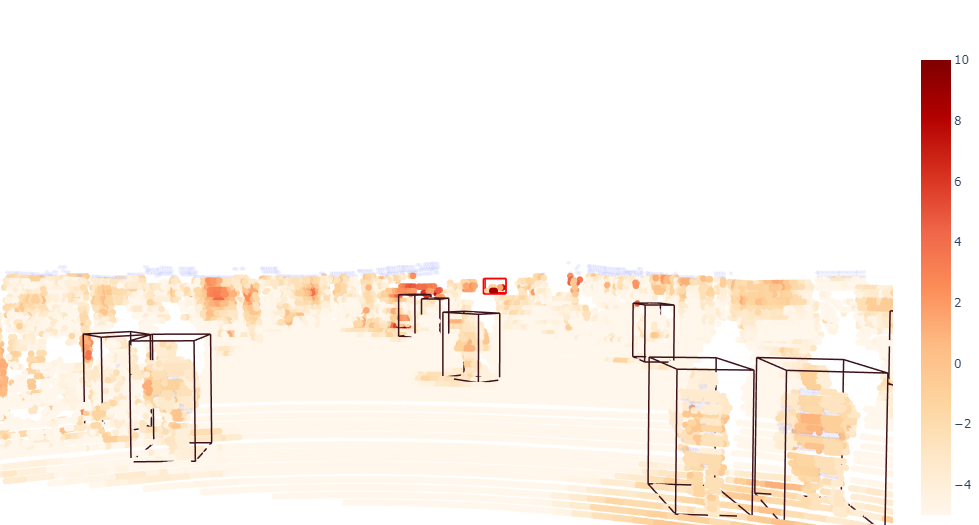}}
			\hspace{-0.3em}
		  \subfigure[]
			{\includegraphics[width=0.48\textwidth, height=0.28\textwidth]{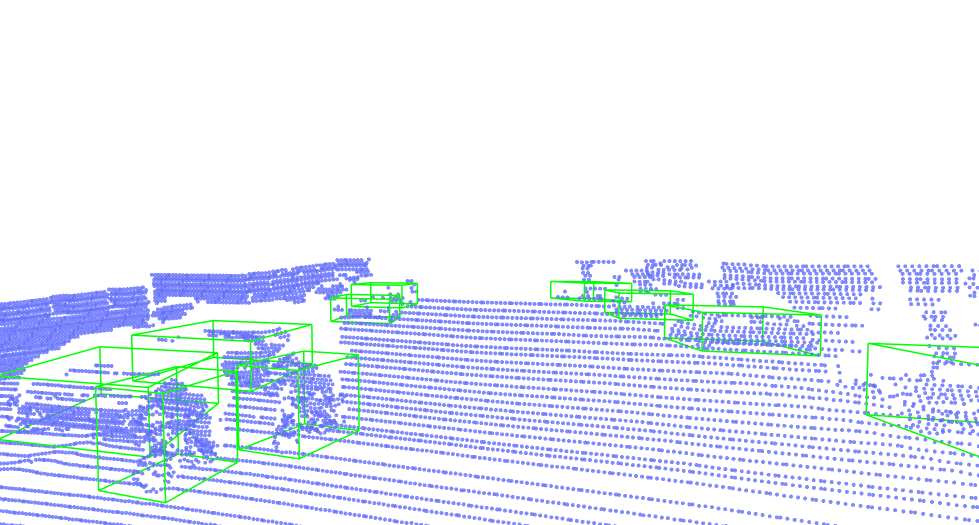}}
		  \subfigure[]
			{\includegraphics[width=0.48\textwidth, height=0.28\textwidth]{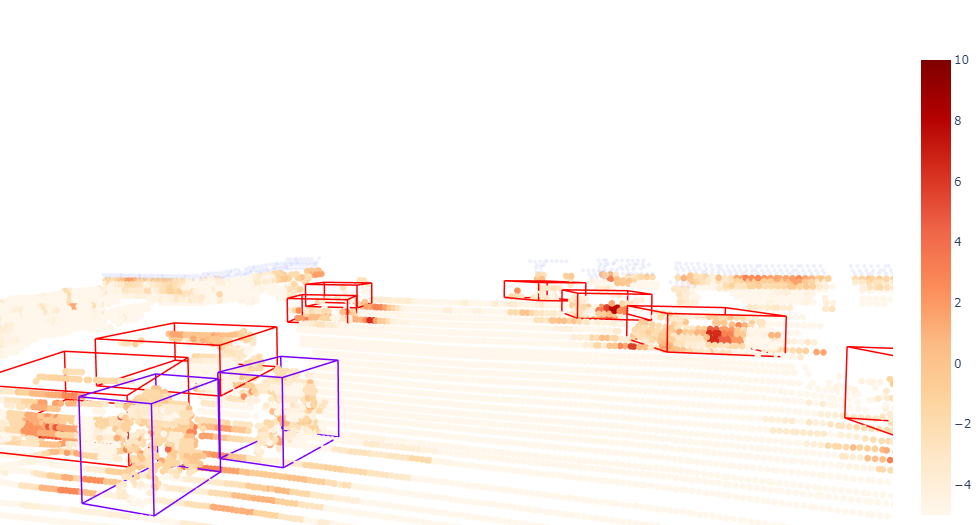}}
			
		% \end{adjustbox}
		\vfill
		\vspace{-0.2cm}
		\caption{Qualitative results on the validation set of KITTI~\citep{geiger2012we}. We show ground-truth bounding boxes as green boxes, while our predictions for the vehicle class are represented with red boxes. We visualize our predictions for the cyclist and pedestrian classes with magenta and violet boxes, respectively. (Best viewed in color.)}
		\label{fig:quali_kitti}    
	  \end{figure*}
	  
	  \subsubsection{Ground truth and feature activations}
	  We show in Fig.~\ref{fig:quali_kitti} input point clouds with ground-truth boxes~({Figs.~\ref{fig:quali_kitti}~(a), (c) and~(e)}) and scene feature activations from the corresponding scenes~({Figs.~\ref{fig:quali_kitti}~(b), (d) and~(f)}) with predicted object boxes. We evaluate our full model with the memory module on the validation set of KITTI dataset. The color bars in the figure represent the level of feature activations, computed as the norm of each scene feature and weighted by the matching probabilities~{W} in Eq.~\ref{eq:wk}. We can observe that the activations of the foreground objects,~\ie activations within the predicted boxes, are higher than those of the surrounding backgrounds~(\eg roads and sidewalks). This indicates that our model focuses more on the foreground objects than the backgrounds, which is crucial for accurate 3D object detection. We can also see that our model mainly focuses on capturing the overall shape of the objects. For example, for the cars in Fig.~\ref{fig:quali_kitti}~(b), it captures the shape of the rear/front part, the window frames, and the wheel wells, while particularly focusing on the bumpers. For pedestrians and cyclists in Figs.~\ref{fig:quali_kitti}~(d) and~(f), the model considers the overall body shapes more than capturing specific parts. This enhances the model's ability to detect and distinguish objects accurately.

	  \begin{figure*}[t]
		\centering
		\captionsetup[subfigure]{aboveskip=1pt,belowskip=1pt,justification=centering}
		% \captionsetup{font={footnotesize}}
		% \begin{adjustbox}{width=1.0\linewidth,center} 
		  \subfigure[]
			{\includegraphics[width=0.7\textwidth, height=0.25\textwidth]{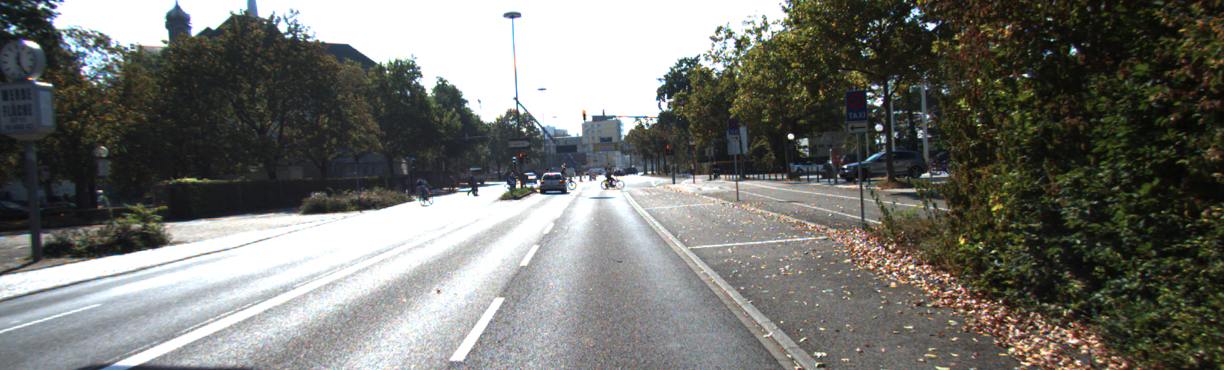}}
			\hspace{-0.3em}
		  \subfigure[]
			{\includegraphics[width=0.7\textwidth, height=0.25\textwidth]{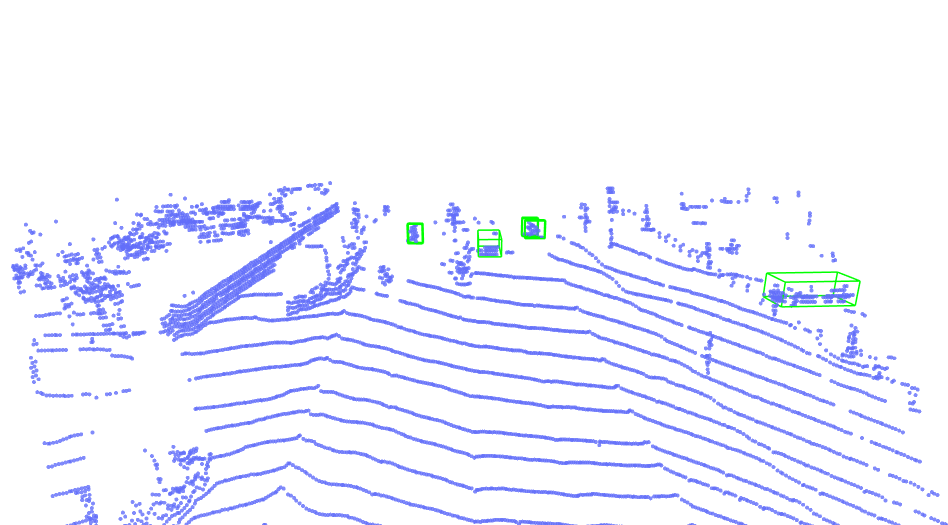}}
			\hspace{-0.3em}
		  \subfigure[]
			{\includegraphics[width=0.7\textwidth, height=0.25\textwidth]{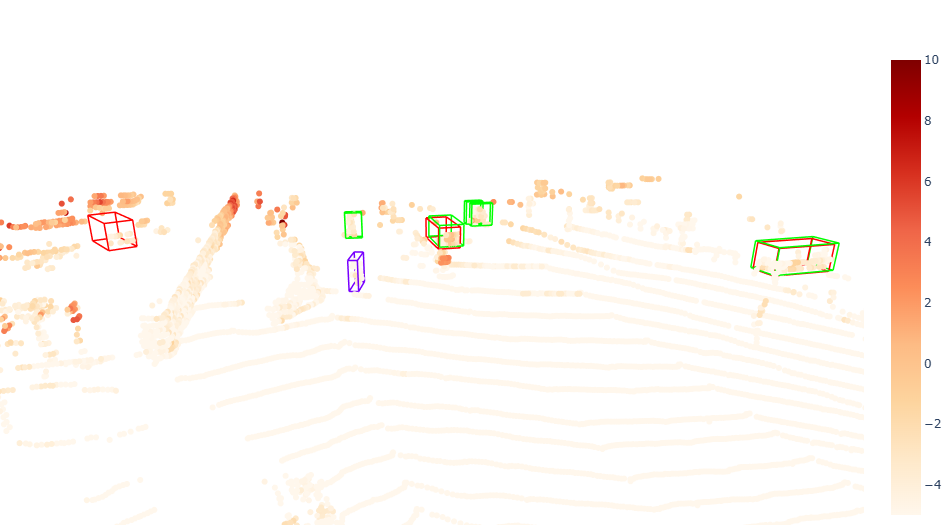}}
		 
		\vfill
		\vspace{-0.2cm}
		\caption{Qualitative results on the validation set of KITTI~\citep{geiger2012we}. We visualize (a) the RGB image, (b) the input point cloud with ground-truth bounding boxes, and (c) the scene feature activations with predicted object boxes. Ground-truth boxes are shown in green, while our predictions for the car and cyclist classes are depicted in red and violet, respectively. Color bars indicate the level of feature activations, computed as the norm of each scene feature. These examples illustrate representative failure cases. (Best viewed in color.)}
		\label{fig:quali_failure}    
	  \end{figure*}

\section{Conclusion and Future Work}

We have presented a novel two-stage 3D detection framework using pseudo image representations. To this end, we have proposed a new CNN architecture, dubbed 3DPillars, based upon the motivation that 3D features from voxels can be seen as a stack of pseudo images. We have implemented this idea using SVFM that enables learning view-specific voxel features without exploiting 3D convolutions. We have also introduced a new RoI head with S$^2$CFM, making it possible to incorporate 3DPillars into a two-stage pipeline, while considering rich scene contexts to refine 3D object proposals. Extensive experiments and analyses demonstrate that 3DPillars achieves a compelling balance between accuracy and efficiency. It offers real-time inference performance (29.6~Hz) with significantly reduced computational cost, while maintaining strong detection results, especially for challenging object categories. Given its modular design, low overhead, and scalability, we believe the proposed framework can serve as a practical and efficient baseline for real-world 3D perception tasks in applications such as autonomous driving, robotics, and smart infrastructure.

\vspace{0.5em}
While the proposed method shows promising performance, it also has several limitations. In Fig.~\ref{fig:quali_failure}, we visualize representative failure cases on the KITTI validation set. As shown in Figs.~\ref{fig:quali_failure}~(a)--(c), our model fails to detect two distant objects—a pedestrian and a cyclist—due to the extremely sparse point cloud representations of those objects. These targets, captured with only a handful of points, lack sufficient structural detail for the model to accurately localize and classify them. Additionally, a false positive occurs where the model incorrectly predicts a car in a garden area, misled by local geometric structures that resemble vehicle parts, highlighting the model's sensitivity to context ambiguity in complex scenes.

These failure cases reflect several intrinsic limitations of the current framework: (1) the difficulty in detecting small or distant objects under extremely sparse input conditions; (2) the inability to leverage semantic cues such as color or texture, which are available in RGB images; and (3) the lack of temporal consistency, which could otherwise help disambiguate dynamic objects like pedestrians and cyclists over time.

\vspace{0.5em}
To overcome these limitations, future work could explore integrating multi-modal inputs (e.g., fusing LiDAR with RGB images) and leveraging temporal information from sequential frames. Such enhancements are expected to improve robustness and reduce both false negatives and false positives. Furthermore, since some false alarms appear to result from overconfident predictions, incorporating confidence calibration techniques~\citep{guo2017calibration} may improve the reliability and interpretability of the model's outputs. These directions offer promising opportunities for advancing both the accuracy and trustworthiness of LiDAR-based 3D object detection systems.

	   \section*{Acknowledgments}
	   This work was supported by Institute of Information \& Communications Technology Planning \& Evaluation~(IITP) grants funded by the Korea government~(MSIT) (No.RS-2022-00143524, Development of Fundamental Technology and Integrated Solution for Next-Generation Automatic Artificial Intelligence System, No.2022-0-00124, Development of Artificial Intelligence Technology for Self-Improving Competency-Aware Learning Capabilities).
	   \vspace{+0.8cm}

% To print the credit authorship contribution details
% \printcredits

%% Loading bibliography style file
% \bibliographystyle{model1-num-names}
% \bibliographystyle{cas-model2-names}
\bibliographystyle{apalike2}
% Loading bibliography database
\bibliography{main} % name your BibTeX data base

\end{document}